\documentclass[lettersize,journal]{IEEEtran}
\usepackage{amsmath,amsfonts}
\usepackage{algorithmic}
\usepackage{algorithm}
\usepackage{array}
\usepackage[caption=false,font=normalsize,labelfont=sf,textfont=sf]{subfig}
\usepackage{textcomp}
\usepackage{stfloats}
\usepackage{url}
\usepackage{verbatim}
\usepackage{graphicx}
\usepackage{cite}
\usepackage{amssymb}
\usepackage{multirow}
\usepackage{adjustbox}
\usepackage{booktabs}
\usepackage{pifont}
\usepackage{hyperref}
\allowdisplaybreaks
\usepackage{tikz,xcolor}

\hypersetup{hidelinks,
	colorlinks=true,
	allcolors=black,
	pdfstartview=Fit,
	breaklinks=true}
	
\definecolor{lime}{HTML}{A6CE39}
\DeclareRobustCommand{\orcidicon}{
\begin{tikzpicture}
\draw[lime, fill=lime] (0,0)
circle[radius=0.16]
node[white]{{\fontfamily{qag}\selectfont \tiny \.{I}D}};
\end{tikzpicture}
\hspace{-2mm}
}
\foreach \x in {A, ..., Z}{%
\expandafter\xdef\csname orcid\x\endcsname{\noexpand\href{https://orcid.org/\csname orcidauthor\x\endcsname}{\noexpand\orcidicon}}
}

\begin{document}

\title{Reliable Multimodal Learning Via Multi-Level Adaptive DeConfusion}

\author{Tong Zhang\hspace{-1.5mm}\orcidA{},~\IEEEmembership{Senior Member,~IEEE,} Shu Shen\hspace{-1.5mm}\orcidB{},~\IEEEmembership{Student Member,~IEEE,} Haiqi Liu\hspace{-1.5mm}\orcidC{},~\IEEEmembership{Student Member,~IEEE,} and C. L. Philip Chen\IEEEauthorrefmark{1}\thanks{* Corresponding author: C. L. Philip Chen.}\hspace{-1.5mm}\orcidD{},~\IEEEmembership{Life Fellow,~IEEE}
\thanks{This work was funded in part by the National Natural Science Foundation of China grant under number 62222603, in part by the STI2030-Major Projects grant from the Ministry of Science and Technology of the People’s Republic of China under number 2021ZD0200700, in part by the Key-Area Research and Development Program of Guangdong Province under number 2023B0303030001, in part by the Program for Guangdong Introducing Innovative and Entrepreneurial Teams (2019ZT08X214), and in part by the Science and Technology Program of Guangzhou under number 2024A04J6310.}
\thanks{The authors are with the Guangdong Provincial Key Laboratory of Computational AI Models and Cognitive Intelligence, the School of Computer Science and Engineering, South China University of Technology, Guangzhou 510006, China, and is with the Pazhou Lab, Guangzhou 510335, China, and is with Engineering Research Center of the Ministry of Education on Health Intelligent Perception and Paralleled Digital-Human, Guangzhou, China. (e-mail: Philip.Chen@ieee.org).}
}

\markboth{Journal of \LaTeX\ Class Files,~Vol.~14, No.~8, January~2025}%
{Shell \MakeLowercase{\textit{et al.}}: A Sample Article Using IEEEtran.cls for IEEE Journals}


\maketitle

\begin{abstract}

Multimodal learning enhances the performance of various machine learning tasks by leveraging complementary information across different modalities. However, existing methods often learn multimodal representations that retain substantial inter-class confusion, making it difficult to achieve high-confidence predictions, particularly in real-world scenarios with low-quality or noisy data. To address this challenge, we propose Multi-Level Adaptive DeConfusion (MLAD), which eliminates inter-class confusion in multimodal data at both global and sample levels, significantly enhancing the classification reliability of multimodal models. Specifically, MLAD first learns class-wise latent distributions with global-level confusion removed via dynamic-exit modality encoders that adapt to the varying discrimination difficulty of each class and a cross-class residual reconstruction mechanism. Subsequently, MLAD further removes sample-specific confusion through sample-adaptive cross-modality rectification guided by confusion-free modality priors. These priors are constructed from low-confusion modality features, identified by evaluating feature confusion using the learned class-wise latent distributions and selecting those with low confusion via a Gaussian mixture model. Experiments demonstrate that MLAD outperforms state-of-the-art methods across multiple benchmarks and exhibits superior reliability.

\end{abstract}


\section{Introduction}
\label{sec:intro}

\begin{figure}[t]
    \centering
    \subfloat[State-of-the-art methods such as MD \cite{han2022multimodal} and PDF \cite{cao2024predictive} tend to produce confused predictions, especially under noisy conditions.]{
        \includegraphics[width=0.95\columnwidth]{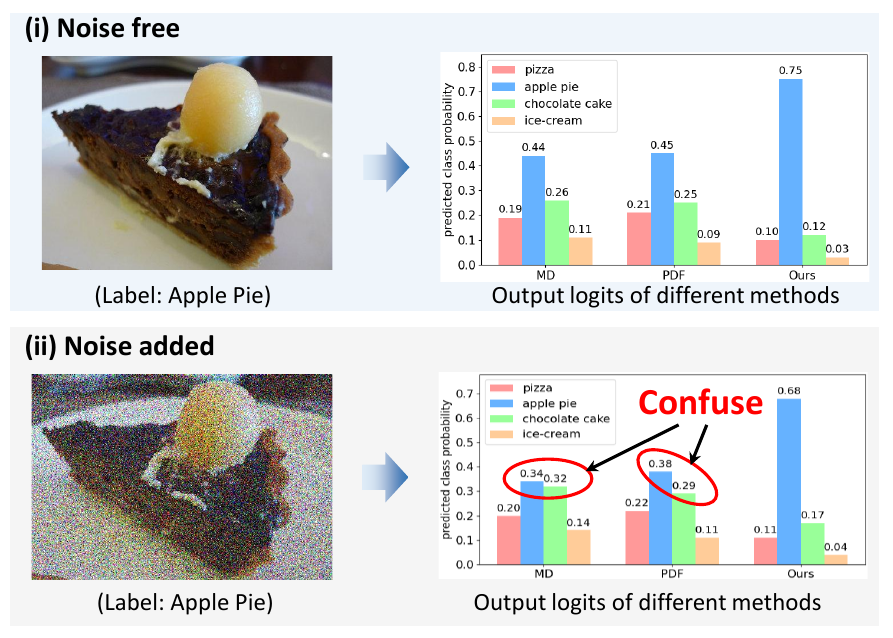}
        \label{fig:motivation-a}
    }
    \vfill
    \subfloat[Illustration of global- and sample-level inter-class confusion.]{
        \includegraphics[width=0.95\columnwidth]{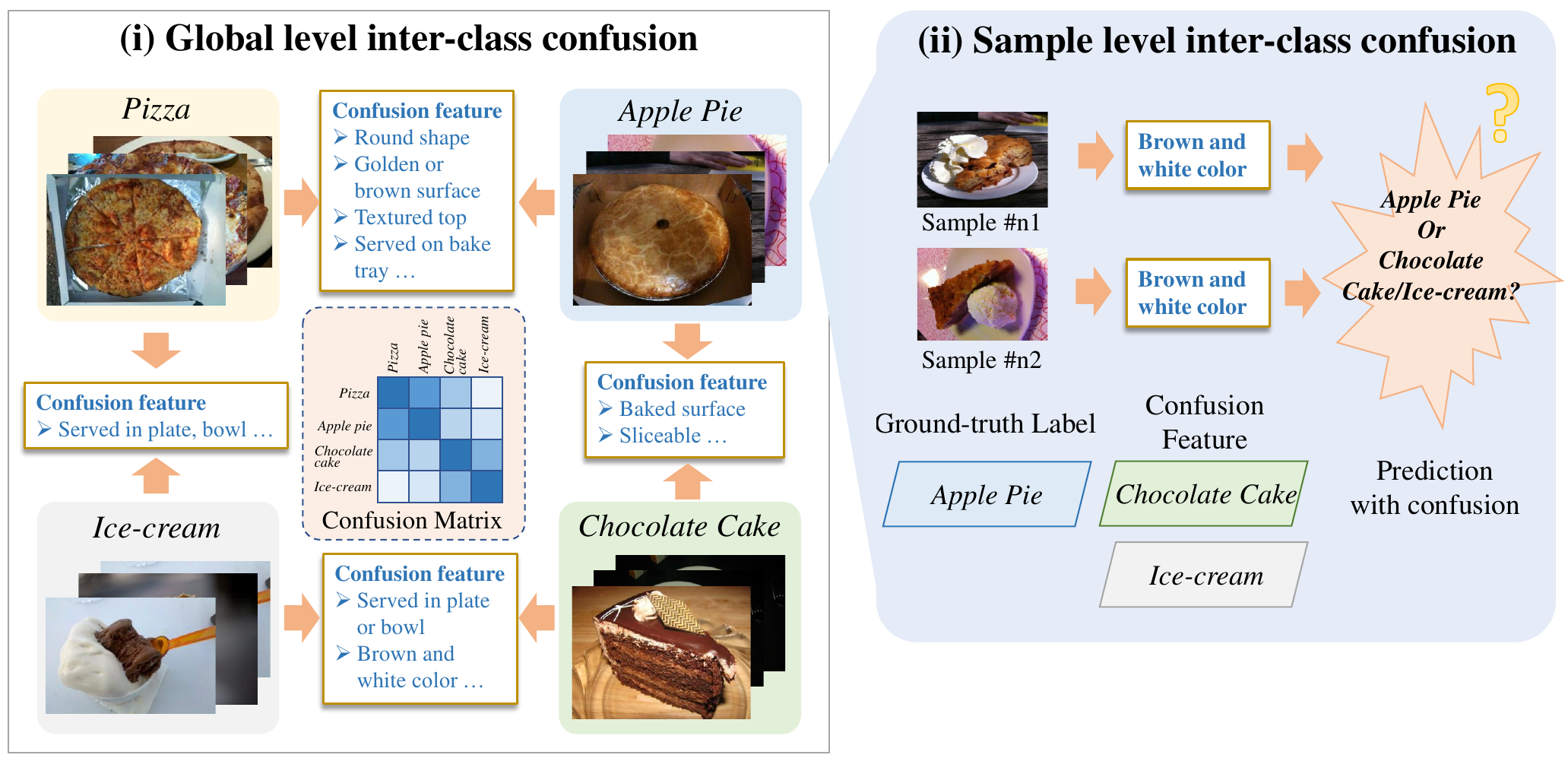}
        \label{fig:motivation-b}
    }
    \caption{Illustration of the motivation of this work. For illustrative purposes, these figures show results on a toy FOOD101 subset containing four classes.}
    \label{fig:motivation}
\end{figure}


\IEEEPARstart{H}{umans} interpret and interact with their environment by combining information from multiple senses, such as sight, sound, and touch \cite{gazzaniga2006cognitive,herreras2010cognitive}. Inspired by this, multimodal learning \cite{dou2022empirical,shankar2018review,liang2022foundations,baltruvsaitis2018multimodal}, which extracts and fuses data from different sensory sources, has received significant attention and achieved remarkable progress. In recent years, it has substantially improved the performance of machine learning across various applications, including medical diagnosis \cite{han2022multimodal,zou2023dpnet,10834580}, image-text classification \cite{zhang2023provable,cao2024predictive,radford2021learning,9444850}, action recognition \cite{kazakos2019epic,gao2020listen}, emotion recognition \cite{hazarika2020misa,sun2022cubemlp}, and audio-visual speech recognition \cite{potamianos2004audio}, among others.

However, in real-world scenarios, multimodal data exhibit highly variable quality. On the one hand, different modalities contain inherently different amounts of task-relevant information. On the other hand, factors such as sensor failures and environmental influences introduce additional noise into the collected data. Despite the rapid development of multimodal methods, they often struggle to learn robust representations and produce high-confidence predictions in these challenging conditions. Many conventional multimodal models lack explicit modeling of input quality and uncertainty, limiting their ability to conduct effective feature enhancement, robust denoising, and reliable fusion. As a result, their performance typically deteriorates significantly as data quality decreases. To address this issue, recent studies have sought to enhance the reliability of multimodal learning through quality-aware feature learning and fusion strategies, thereby substantially reducing the adverse effects of low-quality data. Geng et al. \cite{geng2021uncertainty} have proposed an uncertainty-aware approach to learn noise-free multimodal joint representations. Han et al. \cite{han2022multimodal} have achieved more trustworthy classification results by extracting informativeness features at the feature level and performing dynamic fusion at the modality level. Zhang et al. \cite{zhang2023provable} and Cao et al. \cite{cao2024predictive} have provided theoretical support for reliable multimodal fusion. Zhou et al. \cite{zhou2023calm} have proposed a confidence-based method for feature enhancement and reliable fusion.

However, we find that these methods commonly learn features containing substantial inter-class confusion, which are class-relevant but insufficiently discriminative. As a result, when confronted with low-quality data in which a large portion of discriminative features is corrupted, the models tend to over-rely on such relevant yet indiscriminative features for classification, leading to low-confidence predictions. As shown in Fig. \ref{fig:motivation-a}, existing state-of-the-art methods produce output logits for the illustrated apple pie sample that exhibit high confusion between the apple pie and chocolate cake classes, and this confusion further increases in the presence of noise. We consider the reason could be that these methods learn the brown and white colors of the apple pie, which are class-relevant but insufficiently discriminative features, as such colors are also commonly found in chocolate cake. Under noise-free conditions, these methods can rely on more discriminative fine-grained features of apple pie, such as cross-section texture, to make higher confidence predictions. However, once these discriminative features are corrupted by noise, the methods overly depend on the insufficiently discriminative features, resulting in low-confidence and confused outputs. Therefore, we raise an important question: \textbf{\textit{how to learn confusion-free multimodal representations to enhance the reliability of multimodal learning?}}


To address this issue, this paper proposes Multi-Level Adaptive DeCofusion (MLAD) for reliable multimodal learning. Specifically, we first observe that inter-class confusion manifests at both global and sample levels, and design corresponding adaptive mechanisms for each level to effectively remove the diverse confusion patterns. As shown in Fig. \ref{fig:motivation-b}, \textbf{at the global level}, each class as a whole shares different common features with other classes, and exhibits varying levels of differentiation difficulty. For instance, apple pie and pizza both often have a golden or brown surface and are often served on a baking tray, while apple pie and chocolate cake share other common characteristics. Moreover, the confusion matrix reveals that different classes exhibit varying levels of discrimination difficulties. For instance, the ice-cream class shows relatively low confusion and difficulty in separation from the other three classes, whereas the apple pie class exhibits higher confusion and greater separation difficulty. \textbf{At the sample level}, individual samples also contain unique inter-class confusion that may deviate from global patterns. For example, some of the apple pie samples exhibit the typical appearance, whereas some closely resemble chocolate cake.


Consequently, \textbf{at the global level}, MLAD introduces class-adaptive deconfusion (CAD), which first employs dynamic-exit modality encoders to learn reliable representations and latent distribution for classes of varying discrimination difficulty. The residual cross-class reconstruction is introduced to effectively remove confusion from the learned representations and distributions. Specifically, the dynamic-exit encoder adaptively provides sufficient network depth to capture fine-grained information for hard-to-distinguish classes while avoiding over-fitting to noise on easily separable classes, enabling reliable representation learning. Subsequently, the learned representations for each class are decoupled to compute the residual information remaining in their respective inputs. The residual from each class is then passed through the decoder, and its reconstruction loss with other classes’ inputs is minimized. In this way, MLAD preserves as much information from other classes as possible within the residual information of each class, ensuring that the representation and distribution learned by the modality encoder is discriminative and effectively deconfused. \textbf{At the sample level}, MLAD proposes sample-adaptive deconfusion (SAD), which first uses the class-wise latent distributions learned by CAD to assess the inter-class confusion of modality features with global-level confusion removed. A two-component Gaussian mixture model is then employed to select low-confusion features for constructing confusion-free priors of each modality. Guided by these priors, sample-adaptive cross-modality rectification is applied to remove sample-specific confusion from each sample’s modality features using information from other modalities.

The contributions of this paper can be summarized as:
\begin{itemize}
    \item We observe that existing multimodal methods often learn features that are class-relevant yet insufficiently discriminative, resulting in low confidence. To address this, we propose Multi-Level Adaptive Deconfusion (MLAD) for reliable multimodal learning. Experiments on multiple benchmarks demonstrate the superiority of MLAD over state-of-the-art reliable multimodal learning methods.
    \item Class-adaptive deconfusion (CAD) is introduced in MLAD to remove global-level confusion. By adapting the encoder output depth according to the class-specific discrimination difficulty and employing residual cross-class reconstruction, it effectively eliminates inter-class confusion at the global level.
    \item Sample-adaptive deconfusion (SAD) is proposed in MLAD to further remove sample-level confusion. Guided by confusion-free modality priors, it performs sample-adaptive cross-modality rectification, effectively eliminating sample-specific confusion information, thereby improving classification confidence and robustness.
\end{itemize}

\section{Related Works}
This section briefly reviews some related works on multimodal learning, reliable learning from low-quality multimodal data, and modality quality estimation.
\subsection{Multimodal Learning}
\label{sec:relative-mml}

The advancement of sensor technology and the development of data transmission methods have led to the emergence of a vast amount of data from diverse sources, types, and formats. This has driven the rapid growth of multimodal learning \cite{ramachandram2017deep,baltruvsaitis2018multimodal,zhu2024vision+}, making it a highly focused research hotspot. On the one hand, many multimodal learning methods \cite{natarajan2012multimodal,simonyan2014two,subedar2019uncertainty,arevalo2017gated,hong2020more,hu2021unit,kiela2019supervised,poria2015deep} have significantly enhanced the performance of various applications and tasks. On the other hand, many studies \cite{10089190,9097411,NEURIPS2021_5aa3405a} have focused on a theoretical perspective or drawn inspiration from the brain's processing mechanisms for multimodal information, shedding light on the principles and interpretability of multimodal learning. In recent years, numerous approaches have also targeted the new problems and challenges in multimodal learning brought by real-world scenarios. Multimodal models typically achieve higher accuracy and model reliability compared to unimodal approaches in various applications \cite{10643687,6529074,hang2021multi,huang2020multimodal,10506794,10547435}. For instance, in biomedical applications, DeepIMV \cite{lee2021variational} and MOGONET \cite{wang2021mogonet} have achieved more precise pathological classification by integrating multiple omics data. In emotion recognition tasks, several studies \cite{9863920,10076804,10577436} have proposed instance-adaptive multimodal affective information interaction methods, achieving more robust and generalizable emotion analysis compared to uni-modal methods. In computer vision, CEN \cite{9906429} has enhanced dense image prediction by fusing multimodal data and adaptively exchanging channels of different modality models. Zhou et al. \cite{10122710} have improved motion recognition performance by augmenting multimodal information in videos and mining potential commonality features among modalities. In addition to enhancing various applications, many multimodal studies have also focused on theoretical aspects. For instance, Huang et al. \cite{NEURIPS2021_5aa3405a} have analyzed from a theoretical perspective why multimodal models outperform unimodal models. Some research \cite{10089190,9097411} has explored the mechanistic connections between machine learning models and the human brain by investigating the decoding mechanisms of neural representations. In addition, real-world scenarios pose new issues and challenges for multimodal learning, such as modality imbalance problem \cite{wang2020makes,huang2022modality,wu2022characterizing,fan2023pmr,wei2024diagnosing,10694738} and reliable learning from low-quality data. This paper focuses on reliable multimodal learning in real-world scenarios with noisy and low-quality data, achieving stronger robustness than existing methods by eliminating the inter-class confusion from the learned representations.

\subsection{Reliable Learning from Low-Quality Multimodal Data}
As mentioned in Section \ref{sec:relative-mml}, in real-world scenarios, multimodal data are often noisy, leading to quality degradation. Therefore, how to achieve robust learning on noisy multimodal data has become a widely studied issue. To this end, many studies have developed reliable multimodal learning methods to obtain robust prediction. Federici et al. \cite{federici2020learning} have proposed the multi-view information bottleneck, which has improved the generalization and robustness of multi-view learning by retaining information shared by each view. Han et al. 
 \cite{han2020trusted} have parameterized the evidence of different modality features using Dirichlet distribution and fused each modality at the evidence level using Dempster-Shafer theory. Geng et al. \cite{geng2021uncertainty} have proposed the DUA-Nets, which achieved uncertainty-based multimodal representation learning through reconstruction. Han et al. \cite{han2022multimodal} have modeled informativeness at the feature and modality levels, achieving trustworthy multimodal feature fusion. Zhang et al. \cite{zhang2023provable} have achieved more robust and generalized multimodal fusion by dynamically assigning weights to each modality based on uncertainty estimation. Zheng
et al. \cite{zheng2023multi} have achieved trustworthy multimodal classification via integrating feature and label-level confidence. Zou et al. \cite{zou2023dpnet} have proposed a novel dynamic poly-attention Network that integrated global structural information for trustworthy multimodal classification. Zhou et al. \cite{zhou2023calm} have introduced a trustworthy multi-view classification framework by enhancing multi-view encoding and confidence-aware fusion. Cao et al. \cite{cao2024predictive} have proposed the Predictive Dynamic Fusion method, which provides theoretical guarantees for reducing the upper bound of generalization error in dynamic multimodal fusion. Despite their effectiveness, the representations learned by these methods typically retain a large amount of class-relevant yet insufficiently discriminative features, i.e., inter-class confusion, which leads to highly confused predictions, especially under noisy conditions. To address this issue, we propose Multi-Level Adaptive Deconfusion, which removes inter-class confusion from the learned representations at both global and sample levels, thereby achieving higher classification confidence and improved noise robustness.

\subsection{Learning Representations with Dynamic Network Depth}
In recent years, methods that enable neural networks with dynamic depth to perform adaptive learning and inference for different inputs have been widely explored. Most of these works leverage dynamic-depth networks to improve computational efficiency. For example, Huang et al. \cite{huang2017multi} proposed MSDNet, which places intermediate classifiers at different network depths and designs a computation-aware objective function, allowing easier samples to exit at shallower layers and thereby improving efficiency. Bolukbasi et al. \cite{bolukbasi2017adaptive} dynamically select different network components for each input, enabling examples correctly classified by early layers to exit the network, which also enhances efficiency. Wang et al. \cite{wang2018skipnet} introduced SkipNet, which adaptively skips network blocks for different input samples, achieving sample-specific network depth and significantly reducing computation. Jie et al. \cite{jie2019anytime} proposed Routing Convolutional Network, which determines the optimal output layer for each sample under a given time budget using a time-cost-aware reward function. Other works design dynamic depth mechanisms to mitigate overthinking in deep networks. For instance, Zhou et al. \cite{zhou2020bert} proposed PABEE, allowing the network to exit once a sample produces correct outputs over several consecutive layers. Wang et al. \cite{wang2017idk} introduced the IDK prediction cascades framework to effectively prevent overthinking on easy samples. In this paper, we propose a dynamic-exit modality encoder guided by a residual cross-class reconstruction loss to enable reliable learning for classes of varying discrimination difficulties. Our method is specifically designed for learning robust representations and removing inter-class confusion, and its motivation and design are orthogonal to existing approaches.

\begin{figure*}[t]
\centering
\includegraphics[width=\linewidth]{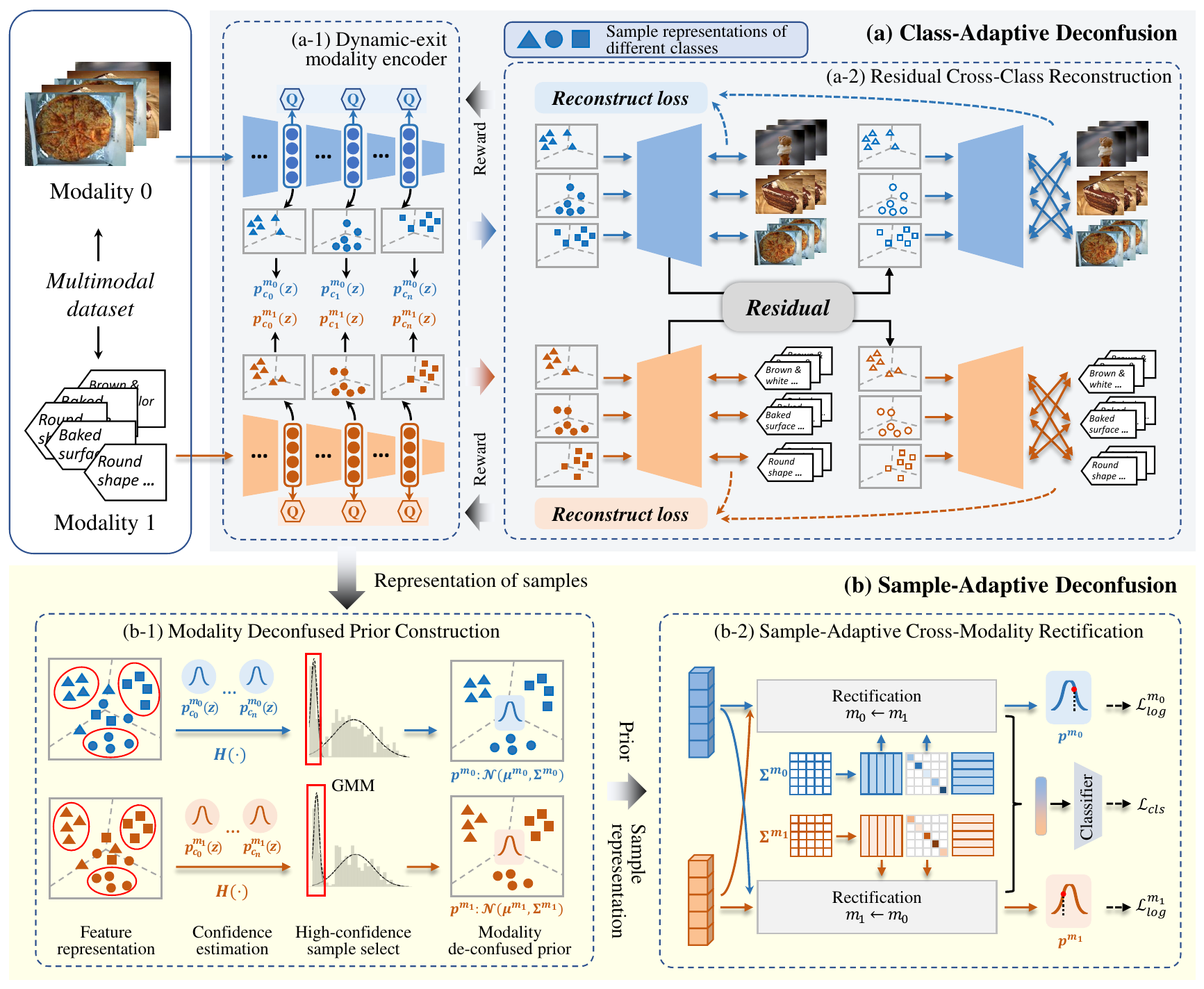} 
\caption{An overview of the proposed MLAD (better viewed in colour). MLAD employs class-adaptive deconfusion and sample-adaptive deconfusion to eliminate global-level confusion and sample-level confusion. Without loss of generality, this figure illustrates the case of two modalities, with blue and orange representing different modalities. The detailed implementation of the Rectification module (gray block) in (b-2) Sample-Adaptive Cross-Modality Rectification is illustrated in Fig. \ref{fig:rec-module}.}
\label{fig:framework_overall}
\end{figure*}

\section{Proposed Approach} \label{sec:prop_frame}
In this section, we present a comprehensive description of the proposed Multi-Level Adaptive DeConfusion (MLAD) framework. Consider a multimodal dataset $\mathcal{D}=\{(x_i,y_i)\}_{i=1}^N$ consisting of $N$ samples, each represented by $M$ modalities and associated with one of $C$ class labels. The goal of multimodal classification is to learn a neural network that maps each input $x_i=\{x_i^m\in\mathbb{R}^{d^m}\}_{i=m}^M$ to its corresponding label $y_i\in\mathbb{R}^C$, where $d^m$ denotes the feature dimension of the $m$-th modality. The proposed MLAD improves the robustness of multimodal classification by adaptively mitigating inter-class confusion across both global and sample levels.

\subsection{Overall Framework of MLAD}
\label{sec:overall-frame}
As shown in Fig. \ref{fig:framework_overall}, the proposed Multi-Level Adaptive Deconfusion (MLAD) eliminates inter-class confusion at the global and the sample levels via Class Adaptive Deconfusion (CAD) and Sample Adaptive Deconfusion (SAD), respectively. CAD focuses on eliminating the confusion information that exists among the overall data distributions of different classes. For a given dataset $\mathcal{D}$, CAD first introduces a modality-specific dynamic-exit encoder to reliably learn representations and distributions of different classes with varying levels of discrimination difficulty within each modality. Subsequently, the residuals between these representations and their corresponding input are computed, and a cross-class reconstruction loss is minimized. In this way, the residuals of each class are enforced to capture all the overlapping information that can be confused with other classes, thereby ensuring that the learned representations and distributions are globally deconfused. Furthermore, variations in sample quality may still cause remaining inter-class confusion that manifests only in specific samples rather than as a global pattern. To this end, SAD is proposed. It first leverages the class-wise distributions to assess the inter-class confusion of each sample representation learned by CAD, and employs a two-component Gaussian mixture model to select low-confusion representations for constructing confusion-free modality priors. Guided by these priors, a sample-adaptive cross-modality rectification is then performed to further eliminate sample-specific confusion from representations with global confusion removed.


\begin{figure}[t]
\centering
\includegraphics[width=\linewidth]{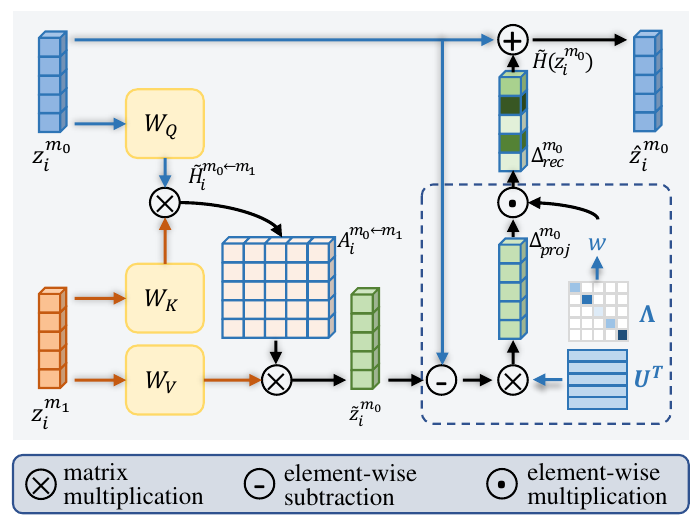} 
\caption{Detailed implementation of Rectification \((m_{0}\leftarrow m_{1})\). Rectification \((m^{1}\leftarrow m^{0})\) is obtained by swapping the input positions of \(z_i^{m_{0}}\) and \(z_i^{m_{1}}\).}
\label{fig:rec-module}
\end{figure}

\subsection{Class Adaptive Deconfusion}\label{sec:cad}
Class-Adaptive Deconfusion (CAD) is first proposed to learn and eliminate the global inter-class confusion information manifested in the overall data distribution. This mechanism consists of two components: the dynamic-exit modality encoder and a residual cross-class reconstruction module. The former is designed to reliably learn the representation distributions of different classes, while the latter thoroughly eliminates the global inter-class confusion information within these representations and distributions.

\subsubsection{Dynamic-Exit Modality Encoder}
As mentioned in Sec. \ref{sec:intro}, considering that the degree of separability between each class and others varies, we first design a dynamic-exit modality encoder that leverages different network depths to learn the representation distributions of different classes. For classes that are difficult to distinguish from others, deeper network layers are employed to capture more fine-grained representations for better discrimination. In contrast, for easily separable classes, excessively deep networks are unnecessary, as they may lead to overfitting to noise. 

Specifically, for network \( \Theta^m \) consisting of \( D^m \) layers of each modality \( m \), an encoder \( \phi^m \) is placed at each layer to encode the intermediate representations and output the latent representation distributions of different classes. The output at the $d$-th layer can be formulated as follows:
\begin{align}
    p_c^m(\textbf{z})\mid _{d}=\phi^m(\Theta^m_{\sim d}(X^m_c))
\end{align}
\( p_c^m(\textbf{z})\mid_{d} \) denotes the latent representation distribution of class \( c \) in modality \( m \) output at the $d$-th layer of network $\Theta^m$. $\Theta_{\sim d}^m$ represents the subnetwork containing the first \( d \) layers of the network \( \Theta^m \). \( X_c^m \) denotes the modality-\( m \) feature inputs of all samples belonging to class \( c \) in the dataset. Therefore, the objective of the Dynamic-Exit Modality Encoder is to determine the best output depth \( d_c^m \) for each class \( c \) in modality $m$, such that the global inter-class confusion information is adequately eliminated. To this end, an objective function \( \mathcal{L}_{\text{CAD}} \) associated with \(\{\{p_c^m(\textbf{z})\}_{c=1}^C\}_{m=1}^m\) is proposed to determine the optimal depths \(\{\{d_c^m\}_{c=1}^C\}_{m=1}^m\), which can be expressed as follows:
\begin{align}
    \{\{d_c^m\}_{c=1}^C\}_{m=1}^m=\mathop{\arg\min}\limits_{\{\{d_c^m\}_{c=1}^C\}_{m=1}^m}\mathcal{L}_{\text{CAD}}(\{\{p_c^m(\textbf{z})\}_{c=1}^C\}_{m=1}^m). \label{eq:cad-obj}
\end{align}
The detailed definition of \( \mathcal{L}_{\text{CAD}} \) and the implementation of optimal depth determination will be elaborated in Sec. \ref{sec:rccr} and \ref{sec:o-cad}.

\subsubsection{Residual Cross-Class Reconstruction} \label{sec:rccr}
To ensure that the inter-class confusion information within the learned representation distributions of each modality and class is thoroughly eliminated, a Residual Cross-Class Reconstruction module is proposed. It enforces that the residuals obtained by excluding the learned representations from each class’s input data contain as much information of other classes as possible.

Specifically, we first introduce \( M \) modality-specific decoders \( \{\psi^m\}_{m=1}^M \). For each modality \( m \), the decoder \( \psi^m \) decodes the resampled class distributions \( p_c^m \) of each class extracted by the Dynamic-Exit Modality Encoder: $\hat{X}_c^m=\psi^m(\textbf{z}_c^m)$. Subsequently, we compute the residuals \( R_c^m \) remaining in the inputs of different classes for each modality, which can be expressed as follows:
\begin{align}
    R_c^m=X_c^m-\hat{X}_c^m+\epsilon,\quad \epsilon\sim\mathcal{N}(0,\sigma_{c,m}^2 I),
    \label{eq:residual-cal}
\end{align}
where $\sigma_{c,m}=\alpha\cdot \mathrm{std}(X_c^m-\psi(\textbf{z}_c^m))$. $\alpha$ is a hyper-parameter and $\mathrm{std}(\cdot)$ computes the standard deviation. Then, we minimize both the reconstruction loss between the encoded features of each class and the corresponding class inputs, and the reconstruction loss between the residuals of each class and all other classes, which can be expressed as follows:
\begin{align}
    \mathcal{L}_{\text{CAD}}&=\frac{1}{MC}\sum_{m=1}^M\sum_{c=1}^C\|\psi^m(\textbf{z}_c^m)-X_c^m\|_{\mathrm{F}}^2 \notag \\
    &+\frac{1}{MC(C-1)}\sum_{m=1}^M\sum_{c=1}^C\sum_{c^{\prime}\neq c}^C\|R_c^m-X_{c^{\prime}}^m\|_{\mathrm{F}}^2 \label{eq:l-re}
\end{align}
Minimizing the first reconstruction loss ensures that the critical information in the inputs is preserved during the encoding process. Minimizing the second reconstruction loss encourages the residuals of each class to retain as much information from other classes as possible, thereby effectively eliminating inter-class confusion from the encoded representations.

\subsubsection{Optimization of CAD} \label{sec:o-cad}
As shown in Eq. (\ref{eq:cad-obj}), CAD aims to determine the optimal network depth \( \{d_c^m\}_{c=1}^C \) for the inputs of different classes across each modality \( m \), such that the loss \( \mathcal{L}_{\text{CAD}} \) is minimized. Since this determination strategy cannot be obtained through conventional backpropagation, inspired by Jie et al. \cite{8936396}, we design a Q-learning–based approach to learn a policy that adaptively determines the output depth according to different class inputs. Specifically, as the input passes through each network layer, the model decides whether to output at the current layer or continue to the next one. This decision-making process is formulated as a Markov Decision Process (MDP), where the feature representation at each layer serves as the MDP state. Given the state at each layer, one of the two actions, $a_{\text{E}}$ or $a_{\text{C}}$, is taken, where $a_{\text{E}}$ indicates outputting at the current layer and $a_{\text{C}}$ indicates proceeding to the next layer. Accordingly, the reward function $r(s,a)$ for taking an action \( a \) given a state \( s \) is defined as follows:
\begin{align}
    r(s,a)=
    \begin{cases}
    \exp(-\mathcal{L}_{\text{CAD}}), & \text{if }a=a_{\text{E}},\\
    0, & \text{if }a=a_{\text{C}}.\\
    \end{cases}
\end{align}
At each layer of the network, a Q-Network is employed to evaluate the Q-values corresponding to different actions under the current state, based on which the optimal output depth policy is derived. During the training phase, we first train the modality networks, the encoders at each layer, and the decoder before training the Q-Network, by minimizing the following loss:
\begin{align}
    \mathcal{L}=\sum_{m=1}^M\sum_{d=1}^{D^m}\mathrm{CE}(\hat{y}_d^m,y)+\mathcal{L}_{\text{CAD}}^*.
\end{align}
\(D^m\) denotes the total number of exits in the \(m\)-th modality network \(\Theta^m\), \(M\) is the number of modalities, \(\hat{y}_d^m\) is the prediction output at the \(d\)-th exit of the \(m\)-th modality network, and $\mathrm{CE}(\cdot)$ denotes the cross-entropy loss. $\mathcal{L}_{\text{CAD}}^*$ is computed using Eq. (\ref{eq:l-re}), where all $z_c^m$ are output by the final layer of the corresponding modality network. Subsequently, in the Q-learning process, each training sample passes through the modality encoder from shallow to deep layers using a \(\epsilon\)-greedy exploration strategy. At each layer, an action is randomly selected between \(a_\text{E}\) and \(a_\text{C}\) with probability \(\epsilon\), while with probability \(1 - \epsilon\), the action corresponding to the higher Q-value predicted by the current layer’s Q-network is chosen. During each Q-learning update, all transitions \((s, a, r, s^{\prime})\) from the entire episode are used as training samples to update the corresponding Q-network. During the inference phase, for each test input, the trained Q-network at each layer evaluates the Q-values of the two actions \(a_\text{E}\) and \(a_\text{C}\) and selects the one with the higher value. If the selected action is \(a_\text{E}\), the output is taken from the current layer. Otherwise, the computation proceeds to the next layer. If all layers select \(a_\text{C}\), the output from the final layer is used.


\subsection{Sample Adaptive Deconfusion}
In Class-Adaptive Deconfusion, the inter-class confusion information exhibiting global-level patterns is eliminated. Sample-Adaptive Deconfusion further removes the inter-class confusion information with sample-specific patterns in each sample. This mechanism consists of two components: confusion-free modality prior construction and sample-adaptive cross-modality rectification.

\subsubsection{Confusion-Free Modality Prior Construction}
For each modality \( m \) input \( x_i^m \) of sample \( x_i \), we first evaluate the degree of inter-class confusion of its representation \( z_i^m \) obtained after Class-Adaptive Deconfusion. Specifically, we use the set of class-wise latent distributions \(\{p_c^m\}_{c=1}^C\) learned in CAD for modality \( m \) to estimate the posterior class probabilities of \( z_i^m \), which can be formulated as $p(c|z_i^m)=p_c^m(z_i^m)$. Subsequently, the degree of inter-class confusion is assessed using the entropy of the posterior probabilities over all classes, which can be expressed as follows:
\begin{align}
    H(z_i^m)=-\sum_{c=1}^C{p(c|z_i^m)\log p(c|z_i^m)}.
\end{align}
The less inter-class confusion present in \( z_i^m \), the more concentrated its class probability distribution and the lower its entropy $H(z_i^m)$. Conversely, higher inter-class confusion corresponds to a more dispersed distribution and higher entropy $H(z_i^m)$. Then, for the entropy \( H=\{H(z_i^m)\}_{i=1}^N \) of modality $m$ of all input samples, we fit a two-component Gaussian Mixture Model, which can be expressed as follows:
\begin{align}
    p(H)=\pi_1\mathcal{N}(H|\mu_1,\sigma_1^2)+\pi_2\mathcal{N}(H|\mu_2,\sigma_2^2),
\end{align}
where $\pi_1$ and $\pi_2$ are mixture coefficients of the two corresponding components, $\mathcal{N}(H|\mu_1,\sigma_1^2)$ and $\mathcal{N}(H|\mu_2,\sigma_2^2)$. The intersection point \( H_{\text{th}} \) of the two Gaussian components, i.e., \( \pi_1\mathcal{N}(H_{th}|\mu_1,\sigma_1^2)=\pi_2\mathcal{N}(H_{th}|\mu_2,\sigma_2^2) \), is computed and used as the threshold to distinguish samples with high inter-class confusion from those with low inter-class confusion, which can be expressed as:
\begin{align}
    \mathcal{S}_{l}^m=\{i_{l}\mid H(z_{i_l}^m)<H_{th}\}, \notag \\
    \mathcal{S}_{h}^m=\{i_{h}\mid H(z_{i_h}^m)\geq H_{th}\}.
\end{align}
Here, \( \mathcal{S}_l^m \) and \( \mathcal{S}_h^m \) denote the sets of sample indices in modality \( m \) corresponding to features with low and high inter-class confusion, respectively. Next, we construct the confusion-free prior distribution \( p^m \) for each modality \( m \) using all features \( \mathcal{Z}_l^m=\{z_{i_l}^m\mid i_l\in \mathcal{S}_l \}\) with low inter-class confusion. Specifically, the distribution \( p^m\sim\mathcal{N}(\mu^m,\Sigma^m) \) is estimated using maximum likelihood estimation, with sample features exhibiting lower inter-class confusion in $\mathcal{Z}_l^m$ assigned higher weights. Therefore, \( \mu^m \) and \( \Sigma^m \) can be expressed as follows:
\begin{align}
    \mu^m=\sum_{i_l=1}^{|\mathcal{S}_l^m|}w_{i_l}^mz_{i_l}^m,\quad \Sigma^m=\sum_{i_l=1}^{|\mathcal{S}_l^m|}w_{i_l}^m(z_{i_l}^m-\mu^m)(z_{i_l}^m-\mu^m)^{\mathrm{T}},
\end{align}
where $w_{i_l}^m=\frac{\exp(-H(z_{i_l}^m))}{\sum_{j=1}^{|\mathcal{S}_{l}^m|}\exp(-H(z_j^m))}$.

\subsubsection{Sample-Adaptive Cross-Modality Rectification}
After obtaining the confusion-free prior distributions \( p^m \) for each modality $m$, we propose a sample-adaptive cross-modality rectification. Guided by the priors \( p^m \), this mechanism removes inter-class confusion information from the representation of each modality in a sample leveraging all other modalities.

Specifically, for an input sample \( x_i \), its modality features \( \{x_i^m\}_{m=1}^M \) are encoded by CAD into representations \( \{z_i^m\}_{m=1}^M \). For each modality \( m \), we use the prior \( p^m \) to design a cross-modality attention mechanism that obtains compensatory information from other modalities \( m^{\prime}(m^{\prime}\neq m, m^{\prime}\in[1,M]) \) to enhance the representation \( z_i^m \). First, the attention scores are computed, which can be expressed as follows:
\begin{align}
    A_i^{m\leftarrow m^{\prime}}=Softmax\left(\frac{(W_Qz_i^m)^\mathrm{T}(W_Kz_i^{m^{\prime}})}{\sqrt{d}}+\Tilde{H}_i^{m\leftarrow m^{\prime}}\right),
    \label{eq:atten-map}
\end{align}
where \( W_Q \) and \( W_K \) are learnable mappings. \( \Tilde{H}_i^{m\leftarrow m^{\prime}} \) denotes the amount of inter-class confusion in modality \( m^{\prime} \) relative to modality \( m \), which can be expressed as follows:
\begin{align}
    \Tilde{H}_i^{m\leftarrow m^{\prime}}=\frac{\exp\left(\frac{H(z_i^m)}{H(z_i^{m^{\prime}})+\epsilon}\right)}{\sum_n^M \exp\left(\frac{H(z_i^m)}{H(z_i^{n})+\epsilon}\right)}.
\end{align}
$\epsilon$ is added to avoid division by zero. Incorporating \( \Tilde{H}_i^{m\leftarrow m^{\prime}} \) into the attention-map computation encourages modality $m$ to draw more compensatory information from other modalities with lower confusion while reducing reliance on those with higher confusion. The compensatory features \( \Tilde{z}_i^m \) obtained by \( z_i^m \) from other modalities \( m^{\prime} \) can be expressed as follows:
\begin{align}
    \Tilde{z}_i^m=\sum_{m^{\prime}\neq m}^M A_i^{m\leftarrow m^{\prime}}(W_Vz_i^{m^{\prime}}).
    \label{eq:z_tilde}
\end{align}
$W_V$ is also a learnable mapping. Before the compensatory features \( \Tilde{z}_i^m \) is added on the modality $m$ representation $z_i^m$, we first reweight it based on the covariance of the prior \( p^m \). Specifically, we first decompose the covariance \( \Sigma^m \) and project the compensatory information onto the principal axes of \( \Sigma^m \):
\begin{align}
    \Sigma^m&=U\Lambda U^{\mathrm{T}},\\
    \Delta_{\mathrm{proj}}^m&=U^{\mathrm{T}}(\Tilde{z}_i^m-z_i^m),
\end{align}
where $U=[u_1,\cdots,u_{d^m}]$, $\Lambda=\mathrm{diag}(\lambda_1,\cdots,\lambda_{d^m})$, $d^m$ is the feature dimensionality of modality $m$. \( \lambda_i \) represents the variance along the corresponding principal axis \( u_i \). Accordingly, we use \( \lambda_1, \ldots, \lambda_{d^m} \) to determine a set of negatively-correlated weights \( w = [w_1, \ldots, w_{d^m}] \), which are computed as follows:
\begin{align}
    w_i=\frac{\exp(-\lambda_i)}{\sum_j^{d^m}\exp(-\lambda_j)}, \quad i\in[1,d^m].
    \label{eq:weight}
\end{align}
Then, the weights \( w \) are multiplied element-wise with \( \Delta_{\mathrm{proj}}^m \) to obtain the reweighted compensatory information:
\begin{align}
    \Delta_{\mathrm{rec}}^m=w\odot \Delta_{\mathrm{proj}}^m.
\end{align}
In the above computation, we decompose the covariance \(\Sigma^{m}\) of the modality-specific prior \(p^{m}\), fitted over samples from all classes, to obtain its principal axes $u_1,\cdots,u_{d^m}$ and their associated variances $\lambda_1,\cdots,\lambda_{d^m}$. A larger variance indicates greater dispersion along that axis, meaning that the corresponding feature values exhibit stronger inter-class differences and thus lower inter-class confusion. Conversely, a smaller variance implies higher inter-class confusion along that direction. Therefore, by assigning weights \(w\) that are negatively correlated with the principal variances \(\lambda\) to the components of the compensatory information projected onto each axis, directions with higher confusion of modality $m$ are encouraged to receive more compensatory information from other modalities. Finally, the reweighted compensatory information is added to rectify the feature \( z_i^m \):
\begin{align}
    \hat{z}_i^m=z_i^m+\Delta_{\mathrm{rec}}^m.
    \label{eq:z_hat}
\end{align}
To train SAD, the negative log-likelihood loss \( \mathcal{L}_{\text{log}} \) and the classification loss \( \mathcal{L}_{\text{cls}} \) are introduced, which are defined as follows:
\begin{align}
    \mathcal{L}_{\text{tot}}&=\mathcal{L}_{\text{log}}+\mathcal{L}_{\text{cls}}\notag \\
    &=\frac{1}{M\times N}\sum_{n=1}^N\sum_{m=1}^M-\log p^m(\hat{z}_i^m)+\mathrm{CE}(\hat{y},y).
\end{align}
\( \mathrm{CE}(\cdot) \) denotes the cross-entropy loss. \( \hat{y} \) represents the predictions obtained by passing the concatenation of all enhanced modality features \( \{\hat{z}_i^m\}_{m=1}^M \) for each sample through the classifier, and \( y \) denotes the ground-truth labels.

\section{Experiments}
In this section, the effectiveness of the proposed MLAD is evaluated on multiple benchmarks by comparing it with seven state-of-the-art reliable multimodal learning methods. In addition, extensive experiments are conducted to verify the effectiveness of each component of MLAD.

\subsection{Experimental Setups} \label{sec:exp-set}

\subsubsection{Datasets}
Experiments are conducted on four commonly used multimodal datasets in previous methods \cite{han2020trusted,geng2021uncertainty,han2022multimodal,zhang2023provable,zheng2023multi,zhou2023calm,pmlr-v235-cao24c}.
\begin{itemize}
    \item \textbf {BRCA}: BRCA \cite{lingle9cancer} is a dataset for breast invasive carcinoma PAM50 subtype classification. The dataset comprises 875 samples, with each sample containing features from three modalities: mRNA expression data (mRNA), DNA methylation data (meth), and miRNA expression data (miRNA). These samples are categorized into five subtypes: Normal-like, Basal-like, HER2-enriched, Luminal A, and Luminal B, with 115, 131, 46, 435, and 147 samples, respectively. BCRA can be obtained from The Cancer Genome Atlas program (TCGA) \footnote{\url{https://www.cancer.gov/aboutnci/organization/ccg/research/structuralgenomics/tcga}}.
    \item \textbf{ROSMAP}: ROSMAP \cite{mukherjee2015religious,a2012overview,de2018multi} is a dataset containing samples from Alzheimer's patients and normal control subjects. The dataset consists of 351 samples, including 182 Alzheimer's disease patients and 169 normal control samples. Each sample includes data from three modalities: mRNA expression data (mRNA), DNA methylation data (meth), and miRNA expression data (miRNA).
    \item \textbf{CUB}: Caltech-UCSD Birds dataset \cite{wah2011caltech} comprises 200 categories of birds. It contains a total of 11,788 samples, with each sample including data from two modalities: images of birds and their corresponding textual descriptions.
    \item \textbf{UPMC FOOD101}: The UPMC FOOD101 dataset \cite{wang2015recipe} comprises food images from 101 categories obtained through Google Image search and corresponding textual descriptions. This dataset contains 90,704 samples, where each sample's image and text are collected from uncontrolled environments, thus inherently containing noise.
\end{itemize}

\begin{table*}[t]
\caption{Comparison with state-of-the-art reliable multimodal classification methods on four datasets.}
\begin{center}
\begin{tabular}{@{}c|ccc|ccc@{}}
\toprule
       \multirow{2}{*}{Method}  & \multicolumn{3}{c|}{BRCA}   & \multicolumn{3}{c}{ROSMAP}    \\ 
 & ACC & WeightedF1 & MacroF1 & ACC & F1 & AUC \\ 
 \midrule 
MD \cite{han2022multimodal} &  87.7$\pm$0.3 & 88.0$\pm$0.5 & 84.5$\pm$0.5       & 84.2$\pm$1.3  & 84.6$\pm$0.7  & 91.2$\pm$0.7       \\ 
MLCLNet \cite{zheng2023multi}& 86.4$\pm$1.6& 87.8$\pm$1.7& 82.6$\pm$1.8      & 84.4$\pm$1.5   & 85.2$\pm$1.5       & 89.3$\pm$1.1  \\ 
DPNET \cite{zou2023dpnet}& 87.8$\pm$1.0& 88.4$\pm$1.2& 85.2$\pm$1.2       & 85.1$\pm$1.1   & 84.8$\pm$0.7         & 91.3$\pm$0.7      \\ 
CALM \cite{zhou2023calm}& 88.2$\pm$0.7& 88.5$\pm$0.8& 85.1$\pm$0.8       & 85.5$\pm$1.2   & 87.9$\pm$0.9         & 91.3$\pm$1.0      \\ 

QMF \cite{zhang2023provable}& 87.4$\pm$0.4& 87.7$\pm$0.5& 84.1$\pm$0.5       & 84.6$\pm$0.9   & 84.8$\pm$0.8         & 90.5$\pm$0.8      \\ 
GCFANet \cite{zheng2024global}& \underline{88.6$\pm$1.5} & \underline{88.9$\pm$1.6} & \underline{85.3$\pm$1.6}     & \underline{86.3$\pm$1.4} & \underline{88.3$\pm$1.6}        & 91.5$\pm$1.2      \\ 
PDF \cite{pmlr-v235-cao24c}& 88.2$\pm$0.7 & 88.0$\pm$0.6 & 84.9$\pm$0.6     & 85.9$\pm$0.5 & 88.0$\pm$0.4        & \underline{91.6$\pm$0.5}  \\
\midrule
MLAD  &\textbf{90.1$\pm$ 0.3} & \textbf{90.2$\pm$ 0.5} & \textbf{88.7$\pm$ 0.4} & \textbf{89.4$\pm$ 0.6} & \textbf{91.6$\pm$ 0.4} & \textbf{93.0$\pm$ 0.4} \\ \bottomrule
\toprule
\multirow{2}{*}{Method}  & \multicolumn{3}{c|}{CUB}   & \multicolumn{3}{c}{FOOD101}    \\  & ACC & WeightedF1 & MacroF1 & ACC & WeightedF1 & MacroF1 \\ \midrule 
MD \cite{han2022multimodal}&  90.1$\pm$0.7 & 90.0$\pm$0.8 & 89.9$\pm$0.7       & 92.8$\pm$0.3  & 92.6$\pm$0.2  & 92.6$\pm$0.2       \\ 
MLCLNet \cite{zheng2023multi}& 88.2$\pm$1.4& 88.3$\pm$1.7& 87.9$\pm$1.3      & 92.1$\pm$1.0   & 92.2$\pm$1.2       & 92.1$\pm$1.1  \\ 
DPNET \cite{zou2023dpnet}& 92.1$\pm$0.9& 91.8$\pm$0.7& 92.2$\pm$1.1       & 93.1$\pm$1.2   & 93.0$\pm$0.7         & 92.8$\pm$0.8      \\ 
CALM \cite{zhou2023calm}& 92.0$\pm$0.2& 92.1$\pm$0.4& 92.1$\pm$0.5       & 93.0$\pm$1.1   & 92.9$\pm$0.7         & 92.9$\pm$0.8      \\ 
QMF \cite{zhang2023provable} & 89.8$\pm$0.4& 89.7$\pm$0.5& 89.1$\pm$0.5       & 92.7$\pm$0.5   & 92.3$\pm$0.4         & 92.4$\pm$0.4      \\ 
GCFANet \cite{zheng2024global}& 92.3$\pm$1.2 & 92.4$\pm$1.1 & 92.6$\pm$1.2     & 92.9$\pm$1.2 & 93.1$\pm$1.1        & 92.7$\pm$0.9      \\ 
PDF \cite{pmlr-v235-cao24c}& \underline{93.0$\pm$0.5} & \underline{93.2$\pm$0.4} & \underline{93.2$\pm$0.4}     & \underline{93.3$\pm$0.6} & \underline{93.5$\pm$0.7}       & \underline{93.0$\pm$0.6}  \\ 
\midrule
MLAD  &\textbf{94.7$\pm$ 0.2} & \textbf{95.1$\pm$ 0.3} & \textbf{95.1$\pm$ 0.5} & \textbf{94.1$\pm$ 0.3} & \textbf{93.8$\pm$ 0.5} & \textbf{94.1$\pm$ 0.5} \\ \bottomrule
\end{tabular}
\end{center}
\label{sota table1}
\end{table*}

\begin{table}[t]
\caption{Comparison with state-of-the-art reliable multimodal classification methods on four datasets when 50\% of the modalities are corrupted with Gaussian noise}
\begin{center}
\begin{adjustbox}{width=0.95\columnwidth}
\begin{tabular}{@{}c|ccc|ccc@{}}
\toprule
       \multirow{2}{*}{Method}  & \multicolumn{3}{c|}{CUB}   & \multicolumn{3}{c}{FOOD101}    \\ 
 & $\sigma=1$ & $\sigma=5$ & $\sigma=10$ & $\sigma=1$ & $\sigma=5$ & $\sigma=10$ \\ 
 \midrule 
MD \cite{han2022multimodal} &  89.2& 69.5 & 62.1       & 90.4  & 72.4  & 70.7       \\ 
MLCLNet \cite{zheng2023multi}& 85.2& 58.6& 39.7      & 89.3   & 65.7       & 47.9  \\ 
DPNET \cite{zou2023dpnet}& 91.1 & 86.4 & 83.6       & 92.2   & 83.5        & 80.4      \\ 
CALM \cite{zhou2023calm}& 91.4& 89.3 & 87.8       & 92.4   & 90.8         & 89.4      \\ 

QMF \cite{zhang2023provable}& 91.5& 89.2 & 87.9      & 92.2   & 90.7         & 89.6      \\ 
GCFANet \cite{zheng2024global}& 92.1& 90.3 & 88.6      & 92.4   & 90.9         & 89.8      \\ 
PDF \cite{pmlr-v235-cao24c}& \underline{92.8}& \underline{91.2} & \underline{89.7}      & \underline{92.5}   & \underline{91.3} &\underline{90.4}  \\
\midrule
MLAD  &\textbf{94.4} & \textbf{93.6} & \textbf{92.4} & \textbf{93.8} & \textbf{92.9} & \textbf{92.3} \\ \bottomrule
\end{tabular}
\end{adjustbox}
\end{center}
\label{tab:guassian-noise}
\end{table}

\begin{table}[t]
\caption{Comparison with state-of-the-art reliable multimodal classification methods on four datasets when 50\% of the modalities are corrupted with Salt-pepper noise}
\begin{center}
\begin{adjustbox}{width=0.95\columnwidth}
\begin{tabular}{@{}c|ccc|ccc@{}}
\toprule
       \multirow{2}{*}{Method}  & \multicolumn{3}{c|}{CUB}   & \multicolumn{3}{c}{FOOD101}    \\ 
 & $\sigma=1$ & $\sigma=5$ & $\sigma=10$ & $\sigma=1$ & $\sigma=5$ & $\sigma=10$ \\ 
 \midrule 
MD \cite{han2022multimodal} &  88.3& 67.6 & 60.5       & 88.5  & 71.5  & 69.7       \\ 
MLCLNet \cite{zheng2023multi}& 83.7& 56.5& 34.2      & 87.7   & 66.1       & 44.8  \\ 
DPNET \cite{zou2023dpnet}& 90.2 & 84.7 & 80.4       & 90.3   & 81.8        & 76.4      \\ 
CALM \cite{zhou2023calm}& 90.3& 87.2 & 85.1       & 91.8   & 90.1         & 88.4      \\ 

QMF \cite{zhang2023provable}& 90.5& 87.1 & 85.4       & 91.4   & 89.7         & 88.1      \\ 
GCFANet \cite{zheng2024global}& 90.7& 87.5 & 84.4       & 92.0   & \underline{90.4}         & 88.7      \\ 
PDF \cite{pmlr-v235-cao24c}& \underline{91.6}& \underline{88.7} & \underline{86.2}      & \underline{92.1}   & \underline{90.4}         & \underline{88.9}  \\
\midrule
MLAD  &\textbf{93.5} & \textbf{91.7} & \textbf{90.8} & \textbf{93.3} & \textbf{91.9} & \textbf{90.7} \\ \bottomrule
\end{tabular}
\end{adjustbox}
\end{center}
\label{tab:pepper-salt}
\end{table}

\subsubsection{Compared Methods}
To demonstrate the improvement in multimodal classification reliability achieved by MLAD, we introduce several representative methods in this domain, including multimodal dynamics (\textbf{MD}) \cite{han2022multimodal}, multi-level confidence learning (\textbf{MLCLNet}) \cite{zheng2023multi}, dynamic poly-attention network (\textbf{DPNET}) \cite{zou2023dpnet}, enhanced encoding and confidence evaluating framework (\textbf{CALM}) \cite{zhou2023calm}, quality-aware multimodal fusion (\textbf{QMF}) \cite{zhang2023provable}, multi-omics data classification network via global and cross-modal feature aggregation (\textbf{GCFANet}) \cite{zheng2024global}, and predictive dynamic fusion (\textbf{PDF}) \cite{pmlr-v235-cao24c}.

\subsubsection{Evaluation Metrics}
\paragraph{BRCA \& CUB \& UMPC FOOD101} BRCA, CUB and UMPC FOOD101 provide multi-class classification task. Three metrics, including accuracy (ACC), average F1 score weighted by support (WeightedF1), and macro-averaged F1 score (MacroF1), are employed to evaluate the performance of different methods.

\paragraph{ROSMAP} ROSMAP provides a binary classification task. The accuracy (ACC), F1 score (F1), and area under the receiver operating characteristic curve (AUC) of different methods are reported by experiment.

\subsubsection{Implementation Details}
\paragraph{Details of Model Implementation} The modality networks in MLAD follow the encoder design of Han et al. \cite{han2022multimodal}. For each modality network, five encoders are stacked, each mapping features to a 128-dimensional latent space. The Q-network at each depth is a single-layer fully connected network, with input dimension equal to the latent dimension and outputting two Q-values corresponding to two action $a_{\text{C}}$ and $a_{\text{E}}$. The decoder uses for residual computation is a fully connected network with a single hidden layer, with input dimension 128, hidden dimension 256, and output dimension equal to the modality input feature dimension. The decoder for cross-class reconstruction is also a fully connected network with one hidden layer, with input and output dimensions equal to the input feature dimension and a hidden dimension of 512.

\paragraph{Experimental Details} Following previous works \cite{geng2021uncertainty,zhang2023provable,zhou2023calm,pmlr-v235-cao24c}, we conduct experiments on data under Gaussian noise to validate the reliability of our method. Following Geng et al. \cite{geng2021uncertainty}, We add noise to half of the data in each modality. Specifically, we generate $\frac{N}{2}$ noise vectors that are sampled from Gaussian distribution $\mathcal{N}(\textbf{0},\textbf{I})$, which can be formulated as $\{\epsilon_i\sim \mathcal{N}(\textbf{0},\textbf{I})\}_{i=1}^{\frac{N}{2}}$. Then, we add these noise vectors multiplied with intensity $\sigma$ to pollute the original data, i.e., $\Tilde{\textbf{x}}_i^m=\textbf{x}_i^m+\sigma\epsilon_i, \forall i\in[1,\frac{N}{2}]$, where $\textbf{x}_i^m$ is the feature vector of the $m$-th modality of the sample with index $i$.
\paragraph{Training Details} We implement the proposed method and other compared methods on the PyTorch 1.12.0 and cuda 11.6 platform, running on Ubuntu 20.04.2 LTS, utilizing one GPU (NVIDIA RTX A6000 with 48 GB of memory) and CPU of AMD EPYC 75F3. The Adam optimizer with learning rate decay is employed to train the model. The initial learning rate of the Adam optimizer is set to 1e-4, the weight decay is set to 1e-4, and the multiplicative factor of the learning rate decay is set to 0.2. All the quantitative results of the proposed MICINet are the average of five random seeds. 

\begin{table*}[]
\caption{Ablation studies on the BRCA, ROSMAP, CUB, and FOOD101 datasets.}
\begin{center}
\begin{tabular}{cccc|ccc|ccc}
\toprule
\multicolumn{4}{c|}{Method}                                                          & \multicolumn{3}{c|}{BRCA}                           & \multicolumn{3}{c}{ROSMAP}    \\ \hline
\multicolumn{1}{c|}{DE} & \multicolumn{1}{c|}{RCCR} & \multicolumn{1}{c|}{CFMP} & CMR & \multicolumn{1}{c|}{$\sigma=0$} & \multicolumn{1}{c|}{$\sigma=5$} & $\sigma=10$ & \multicolumn{1}{c|}{$\sigma=0$} & \multicolumn{1}{c|}{$\sigma=5$} & $\sigma=10$ \\ \hline
\multicolumn{1}{c|}{\ding{55}}  & \multicolumn{1}{c|}{\ding{51}}    & \multicolumn{1}{c|}{\ding{51}}   & \ding{51}   & \multicolumn{1}{c|}{89.4} & \multicolumn{1}{c|}{87.6} & 84.5 & \multicolumn{1}{c|}{89.1} & \multicolumn{1}{c|}{87.3} & 86.7\\ \hline
\multicolumn{1}{c|}{\ding{51}}  & \multicolumn{1}{c|}{\ding{55}}    & \multicolumn{1}{c|}{\ding{51}}   & \ding{51}   & \multicolumn{1}{c|}{85.6} & \multicolumn{1}{c|}{79.2} & 70.9 & \multicolumn{1}{c|}{84.8} & \multicolumn{1}{c|}{76.4} & 71.5\\ \hline
\multicolumn{1}{c|}{\ding{51}}  & \multicolumn{1}{c|}{\ding{51}}    & \multicolumn{1}{c|}{\ding{55}}   & \ding{51}   & \multicolumn{1}{c|}{88.7} & \multicolumn{1}{c|}{86.1} & 83.7 & \multicolumn{1}{c|}{88.7} & \multicolumn{1}{c|}{86.6} & 83.4\\ \hline
\multicolumn{1}{c|}{\ding{51}}  & \multicolumn{1}{c|}{\ding{51}}    & \multicolumn{1}{c|}{\ding{51}}   & \ding{55}   & \multicolumn{1}{c|}{88.2} & \multicolumn{1}{c|}{84.2} & 78.4 & \multicolumn{1}{c|}{88.1} & \multicolumn{1}{c|}{83.7} & 77.2\\ \hline
\multicolumn{1}{c|}{\ding{51}}  & \multicolumn{1}{c|}{\ding{51}}    & \multicolumn{1}{c|}{\ding{51}}   & \ding{51}   & \multicolumn{1}{c|}{\textbf{90.1}} & \multicolumn{1}{c|}{\textbf{88.7}} & \textbf{86.5} & \multicolumn{1}{c|}{\textbf{89.4}} & \multicolumn{1}{c|}{\textbf{88.4}} & \textbf{87.9}\\ 

\bottomrule
\toprule
\multicolumn{4}{c|}{Method}                                                          & \multicolumn{3}{c|}{CUB}                           & \multicolumn{3}{c}{FOOD101}    \\ \hline
\multicolumn{1}{c|}{DE} & \multicolumn{1}{c|}{RCCR} & \multicolumn{1}{c|}{CFMP} & CMR & \multicolumn{1}{c|}{$\sigma=0$} & \multicolumn{1}{c|}{$\sigma=5$} & $\sigma=10$ & \multicolumn{1}{c|}{$\sigma=0$} & \multicolumn{1}{c|}{$\sigma=5$} & $\sigma=10$ \\ \hline
\multicolumn{1}{c|}{\ding{55}}  & \multicolumn{1}{c|}{\ding{51}}    & \multicolumn{1}{c|}{\ding{51}}   & \ding{51}   & \multicolumn{1}{c|}{93.2} & \multicolumn{1}{c|}{89.4} & 86.9 & \multicolumn{1}{c|}{92.7} & \multicolumn{1}{c|}{88.5} & 86.1\\ \hline
\multicolumn{1}{c|}{\ding{51}}  & \multicolumn{1}{c|}{\ding{55}}    & \multicolumn{1}{c|}{\ding{51}}   & \ding{51}   & \multicolumn{1}{c|}{89.4} & \multicolumn{1}{c|}{83.5} & 75.6 & \multicolumn{1}{c|}{89.1} & \multicolumn{1}{c|}{82.6} & 74.2\\ \hline
\multicolumn{1}{c|}{\ding{51}}  & \multicolumn{1}{c|}{\ding{51}}    & \multicolumn{1}{c|}{\ding{55}}   & \ding{51}   & \multicolumn{1}{c|}{92.5} & \multicolumn{1}{c|}{87.7} & 84.9 & \multicolumn{1}{c|}{91.7} & \multicolumn{1}{c|}{86.1} & 83.8\\ \hline
\multicolumn{1}{c|}{\ding{51}}  & \multicolumn{1}{c|}{\ding{51}}    & \multicolumn{1}{c|}{\ding{51}}   & \ding{55}   & \multicolumn{1}{c|}{91.3} & \multicolumn{1}{c|}{85.4} & 81.7 & \multicolumn{1}{c|}{91.1} & \multicolumn{1}{c|}{84.4} & 80.9\\ \hline
\multicolumn{1}{c|}{\ding{51}}  & \multicolumn{1}{c|}{\ding{51}}    & \multicolumn{1}{c|}{\ding{51}}   & \ding{51}   & \multicolumn{1}{c|}{\textbf{94.7}} & \multicolumn{1}{c|}{\textbf{92.6}} & \textbf{91.4} & \multicolumn{1}{c|}{\textbf{94.1}} & \multicolumn{1}{c|}{\textbf{92.9}} & \textbf{92.3}\\ 
\bottomrule
\end{tabular}
\end{center}
\label{tab:ablation}
\end{table*}

\subsection{Comparison With State-of-The-Art Methods}\label{sec:comp-sota}
\paragraph{Without adding noise}Following previous works \cite{geng2021uncertainty,zhou2023calm}, we compare the classification performance of different methods on the original datasets without introducing additional noise. As shown in Table \ref{sota table1}, compared with other reliable multimodal learning methods, the proposed MLAD consistently achieves superior classification performance across all datasets. On the BRCA dataset, MLAD surpasses the state-of-the-art methods by 1.5\%, 1.3\%, and 3.4\% in terms of ACC, Macro-F1, and Weighted-F1, respectively. On the CUB dataset, the improvements are 1.7\%, 1.9\%, and 1.9\% for the same metrics. On the FOOD101 dataset, MLAD achieves 0.8\%, 0.3\%, and 1.1\% gains in ACC, Macro-F1, and Weighted-F1, respectively. On the ROSMAP dataset, the ACC, F1, and AUC scores are improved by 3.1\%, 3.3\%, and 1.5\%, respectively. As illustrated in Fig. \ref{fig:motivation-b}, the original datasets inherently contain inter-class confusion, while MLAD effectively learns and removes such confusion, leading to stronger separability across different classes.
\paragraph{Under noise}We further compare the performance of different methods on datasets with added noise. Following previous work \cite{geng2021uncertainty,zhang2023provable,cao2024predictive}, Gaussian noise and salt-pepper noise are introduced to the image modality. Table \ref{tab:guassian-noise} presents the performance comparison under Gaussian noise. As the noise intensity increases, the performance of all methods decreases to varying degrees. Among the comparison methods, PDF exhibits a relatively slower performance degradation, with accuracy drops of 1.6\% and 3.1\% at noise intensities of $\epsilon = 5$ and $\epsilon = 10$, respectively, compared to the noise-free setting. In contrast, the proposed MLAD demonstrates greater stability, with accuracy drops of only 0.8\% at $\epsilon = 5$ and 2.0\% at $\epsilon = 10$. Table \ref{tab:pepper-salt} shows the performance comparison under salt-pepper noise. MLAD again exhibits a milder performance decline with increasing noise intensity compared to other methods. A possible reason is that MLAD avoids over-reliance on insufficient discriminative class features under noisy conditions, instead acquiring more discriminative representations through cross-modality rectification. Consequently, MLAD achieves higher classification confidence and robustness than other methods.

\subsection{Ablation study}
Comprehensive ablation studies are conducted to validate the contribution of each component in MLAD to improving model performance and robustness. Specifically, we examine the effects of the dynamic-exit mechanism (DE) in the modality encoder, Residual Cross-Class Reconstruction (RCCR), Confusion-Free Modality Prior (CFMP), and sample-adaptive cross-modality rectification (CMR). In the ablated versions, DE is removed by forcing all inputs to output from the deepest encoder layer. RCCR is removed by excluding the second term of \( \mathcal{L}_{\text{CAD}} \) in Eq. (\ref{eq:l-re}). CFMP is removed by replacing \(\Delta_{\text{rec}}^m\) in Eq. (\ref{eq:z_hat}) with \(\Tilde{z}_i^m\) computed from Eq. (\ref{eq:z_tilde}). CMR is removed by directly feeding the class-adaptive deconfusion features into the classifier for prediction.

Table \ref{tab:ablation} compares the performance of different ablation variants under various noise settings. The removal of each component leads to a decline in both performance and noise stability of MLAD, demonstrating the contribution of each module to enhancing model robustness. The degree of performance and stability degradation, from greatest to least, is observed for RCCR, CMR, CFMP, and DE. The relatively large contributions of RCCR and CMR to robustness arise from their critical roles in eliminating global and sample-level inter-class confusion, respectively. Since all class samples are affected by noise, minimizing the reconstruction loss between each class residual and the inputs from other classes in RCCR effectively retains noise information in the residuals, thereby ensuring that the encoded representations are denoised. CMR further removes sample-level noise by perceiving the confidence of each modality and leveraging cross-modality rectification. CFMP enhances model reliability by providing confusion-free priors for CMR, while DE produces more reliable representations guided by RCCR.

\begin{figure*}[t]
    \centering
    \subfloat[Visualization of Global Inter-Class Confusion Information Across Five Categories in the BRCA Dataset]{
        \includegraphics[width=0.95\columnwidth]{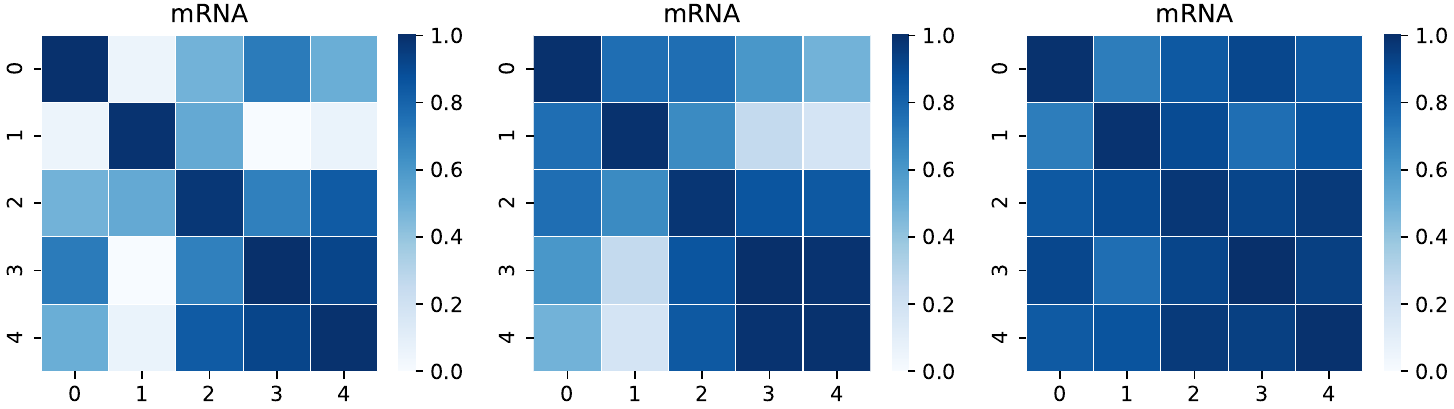}
        \label{fig:class-confusion}
    }
    \hfill
    \subfloat[Output Network Depths of the Dynamic-Exit Modality Encoder for Different Class Inputs on the BRCA Dataset]{
        \includegraphics[width=0.95\columnwidth]{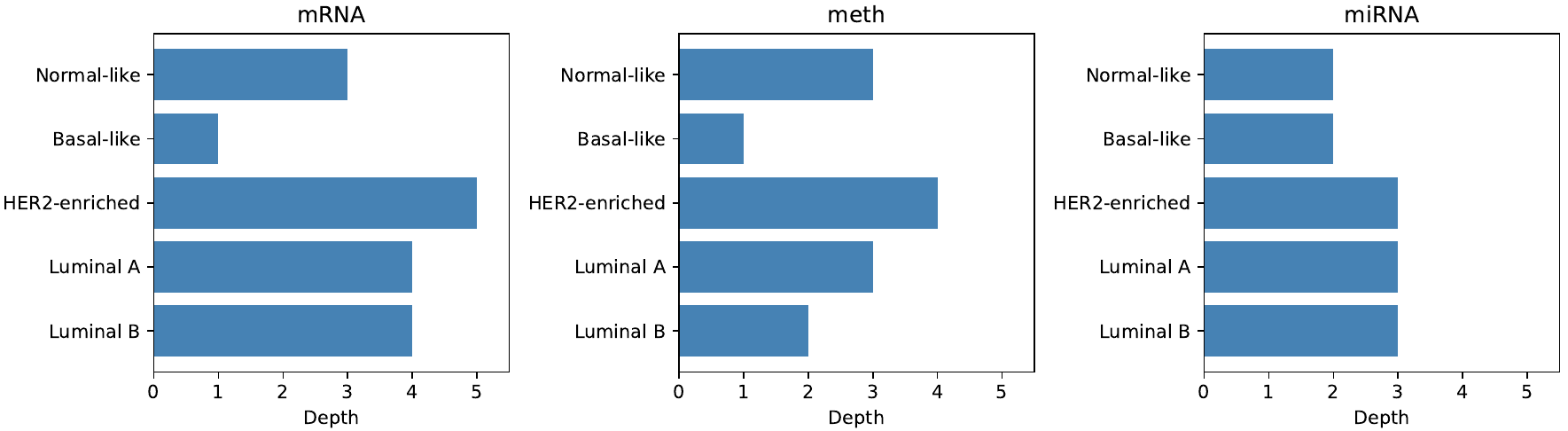}
        \label{fig:vis-dydepth}
    }
    \caption{Illustration of the differences in inter-class separability across modalities and the corresponding output depths determined by the Dynamic-Exit Modality Encoder for different classes. (a) Certain classes exhibit lower overall similarity to other classes (e.g., Class 1 in the mRNA modality), indicating lower discrimination difficulty, whereas others show higher similarity (e.g., Class 2 in mRNA), indicating higher discrimination difficulty. (b) The Dynamic-Exit Modality Encoder adaptively adjusts the output depth according to the discrimination difficulty of each class, enabling classes with lower difficulty to exit from shallower layers, while more complex classes are processed at deeper layers. For example, the Basal-like class (corresponding to Class 1 in (a)) exits at the shallowest layer, whereas the HER2-enriched class (corresponding to Class 2 in (a)) exits at the deepest layer.
}
    \label{fig:tsne}
\end{figure*}

\begin{figure*}[t]
    \centering
    \subfloat[t-SNE visualizations of the original input features from each modality]{
        \includegraphics[width=0.65\columnwidth]{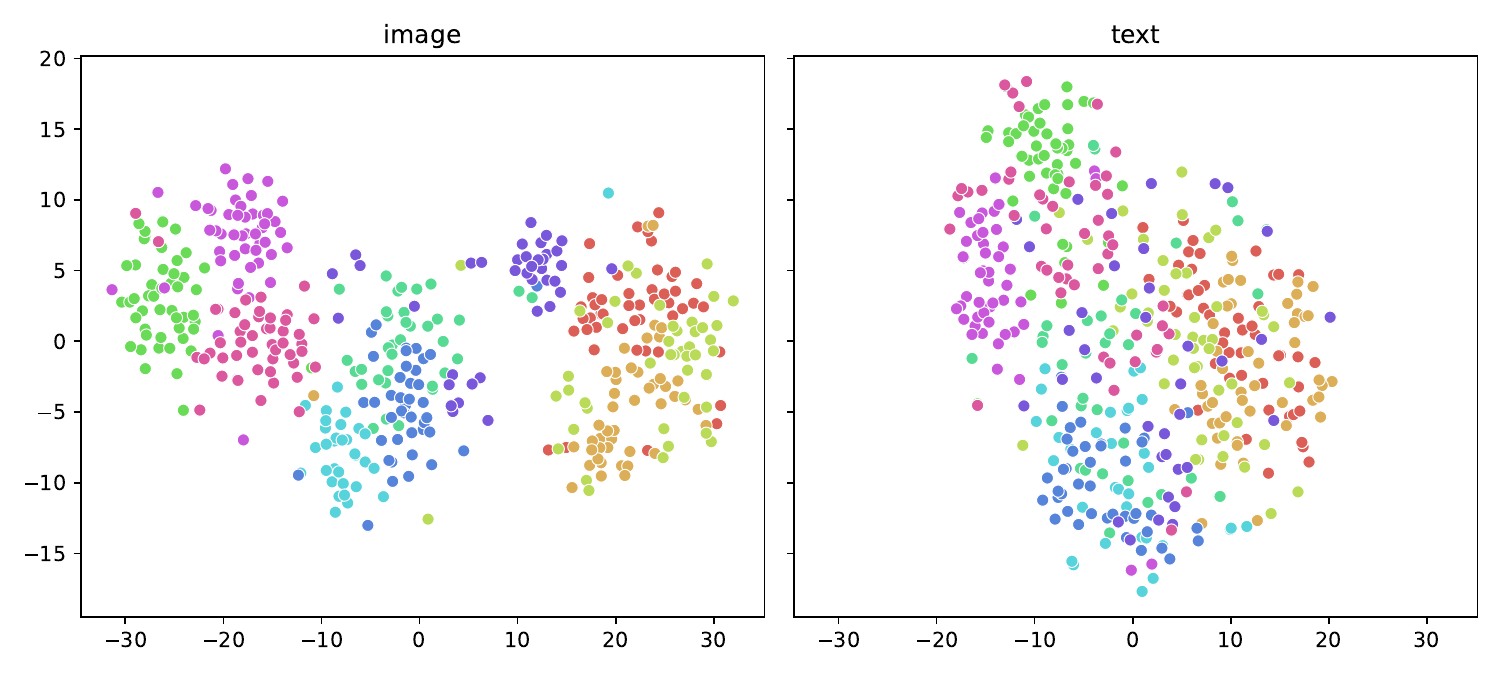}
        \label{fig:tsne-ori}
    }
    \hfill
    \subfloat[t-SNE visualizations of the representations of each modality learned by the encoders without RCCR]{
        \includegraphics[width=0.65\columnwidth]{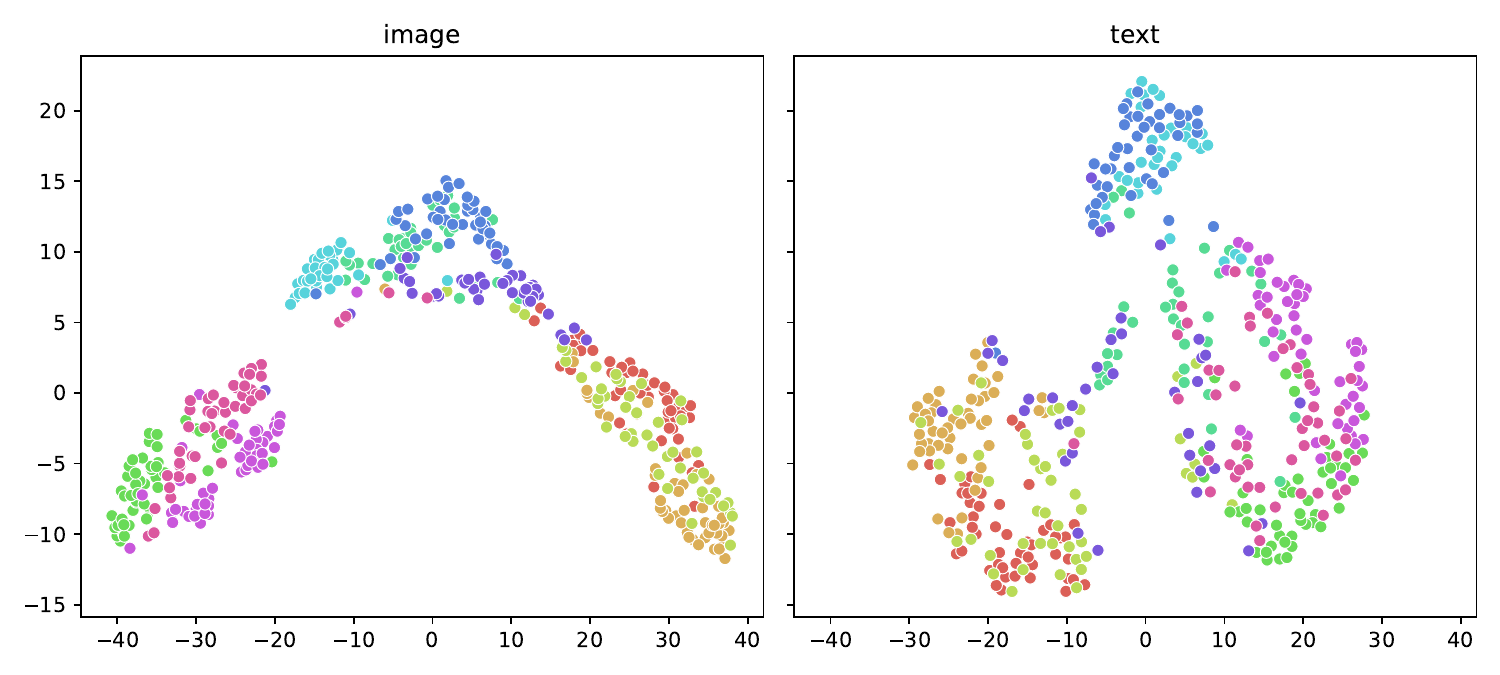}
        \label{fig:tsne-wo-cad}
    }
    \hfill
    \subfloat[t-SNE visualizations of the representations of each modality learned by the encoders with RCCR]{
        \includegraphics[width=0.65\columnwidth]{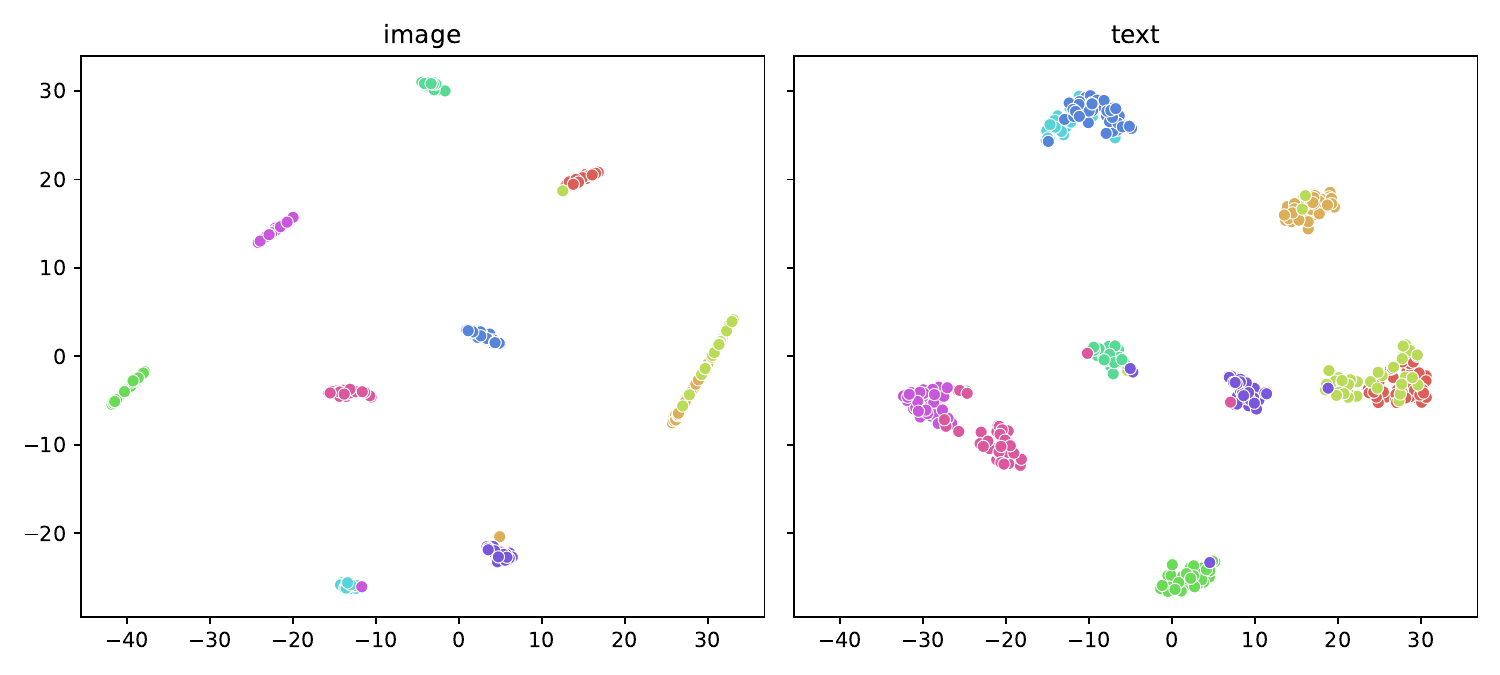}
        \label{fig:tsne-cad}
    }
    \caption{t-SNE visualizations of the original modality features, the learned representations by the encoders without and with RCCR on the CUB dataset.}
    \label{fig:tsne}
\end{figure*}

\begin{table}[t]
\caption{Comparison of classification accuracy under different noise intensities for MLAD and its ``depth-plus-one version'' and ``depth-minus-one version.''}
\begin{center}
\begin{tabular}{c|c|c|c|c}
\hline
Dataset               & Method & $\sigma=0$ & $\sigma=5$ & $\sigma=10$\\ \hline
\multirow{3}{*}{BRCA}   & MLAD(-1)     & 89.4 & 88.3 & 85.8  \\
                        & MLAD(+1)     & 89.5 & 88.0 & 85.6  \\
                    & MLAD      & \textbf{90.1} & \textbf{88.7} & \textbf{86.5}\\ 
\hline

\multirow{3}{*}{ROSMAP}  & MLAD(-1)     & 88.7 & 87.9 & 87.3  \\
                        & MLAD(+1)     & 88.9 & 87.7 & 86.8  \\
                      & MLAD      & \textbf{89.4} & \textbf{88.4} & \textbf{87.9} \\ 
\hline
\multirow{3}{*}{CUB} & MLAD(-1)     & 94.0 & 92.3 & 91.1  \\
                        & MLAD(+1)     & 94.3 & 92.1 & 90.8  \\
                      & MLAD     & \textbf{94.7} & \textbf{92.6}& \textbf{91.4} \\ 
                      \hline

\multirow{3}{*}{FOOD101}  & MLAD(-1)     & 93.4 & 92.5& 91.8  \\
                        & MLAD(+1)     & 93.5 & 92.2& 91.5  \\
                      & MLAD      & \textbf{94.1} & \textbf{92.9} & \textbf{92.3}  \\ 
                      \hline
\end{tabular}
\end{center}
\label{tab:opt-depth}
\end{table}

\subsection{Discussion on class-adaptive deconfusion}
The class-adaptive deconfusion (CAD) module consists of two components: the dynamic-exit modality encoder and residual cross-class reconstruction. Additional experiments are conducted to further investigate the effectiveness of these two components.
\subsubsection{Dynamic-Exit Modality Encoder}
As introduced in Sec. \ref{sec:intro}, different classes exhibit varying levels of discrimination difficulty against other classes. The dynamic-exit modality encoder determines the output depth for each class according to its discrimination difficulty. In Fig. \ref{fig:class-confusion}, we visualize the confusion matrices among different classes in each modality on the BRCA dataset. For each modality, classes with lighter-colored rows and columns indicate lower similarity and confusion with other classes, while darker ones indicate higher confusion. This confirms that the classes in the data indeed present varying discrimination difficulties. Fig. \ref{fig:vis-dydepth} shows the output depths corresponding to each class in the dynamic-exit modality encoder, where Normal-like, Basal-like, HER2-enriched, Luminal A, and Luminal B correspond to Classes 0–4 in Fig. \ref{fig:class-confusion}, respectively. The output depths of each class are well aligned with the discrimination difficulties observed in Fig. \ref{fig:class-confusion}. For example, in the mRNA modality, Class 1 (i.e., Basal-like) exhibits the strongest separability from other classes in Fig. \ref{fig:class-confusion} and correspondingly has the shallowest output depth in Fig. \ref{fig:vis-dydepth}. Conversely, Class 2 (i.e., HER2-enriched) shows the weakest separability and therefore has the deepest output depth. This demonstrates that the dynamic-exit modality encoder can adaptively employ deeper networks for classes with higher discrimination difficulty to capture more detailed information. For easier classes, it uses shallower networks to avoid overfitting to noisy information, thereby enabling more reliable feature representation.

Ablation studies in Table \ref{tab:ablation} have shown that removing the dynamic-exit mechanism and forcing all outputs from the deepest network layer reduces model reliability. Further experiments confirm that the output depths determined by dynamic-exit for each class are optimal. As shown in Table \ref{tab:opt-depth}, under Gaussian noise with $\epsilon=5$, we let the five classes in BRCA output from one layer shallower (-1) or one layer deeper (+1) than the depths suggested by dynamic-exit. Compared with the dynamic-exit–recommended depths (DE), both the -1 and +1 settings lead to a decrease in model performance. This indicates that the output depths assigned by the dynamic-exit mechanism are the most suitable.

\begin{figure*}[t]
\centering
\includegraphics[width=0.95\textwidth]{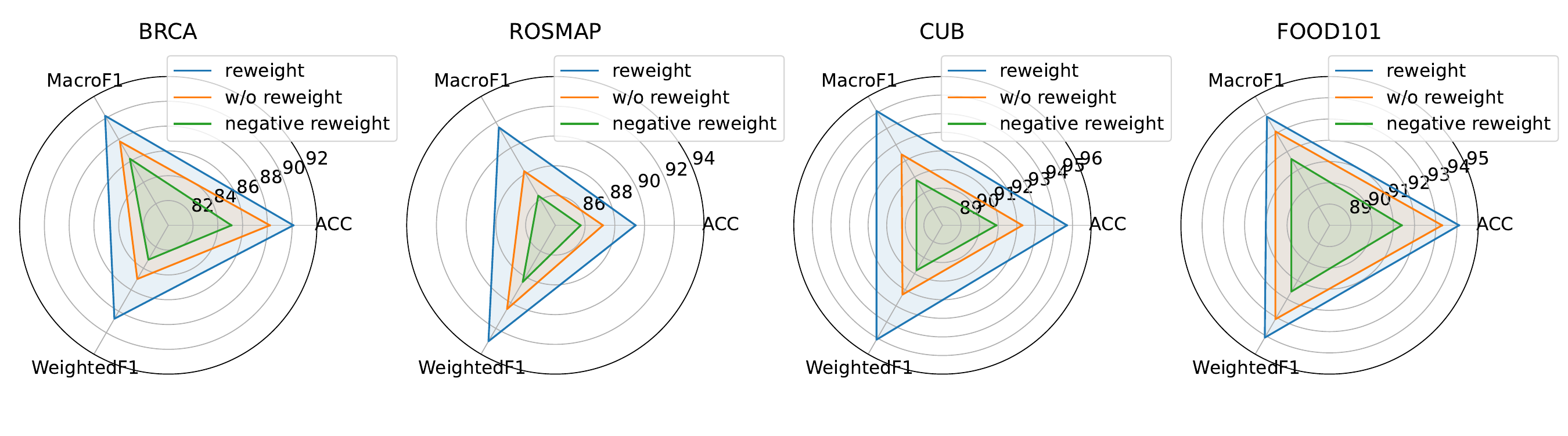} 
\caption{Comparison of MLAD classification performance on four datasets under SAD with cross-modal compensatory information using reweighting, without reweighting, and with negative reweighting.}
\label{fig:reweight}
\end{figure*}

\begin{figure}[t]
    \centering
    \subfloat[Result on dataset without adding noise.]{
        \includegraphics[width=0.95\columnwidth]{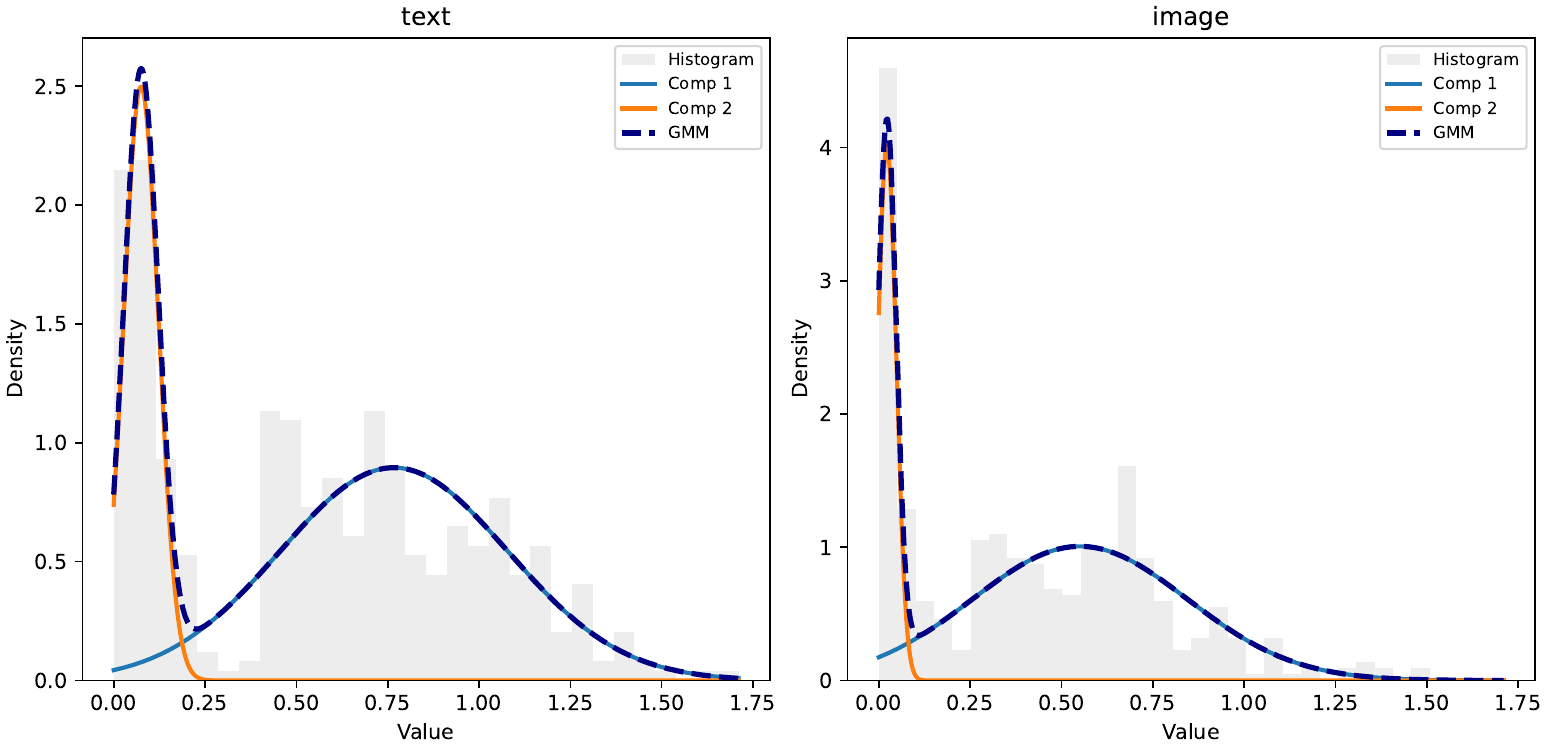}
        \label{fig:gmm-noise-free}
    }
    \hfill
    \subfloat[Result on dataset under Guassian noise ($\sigma=5$).]{
        \includegraphics[width=0.95\columnwidth]{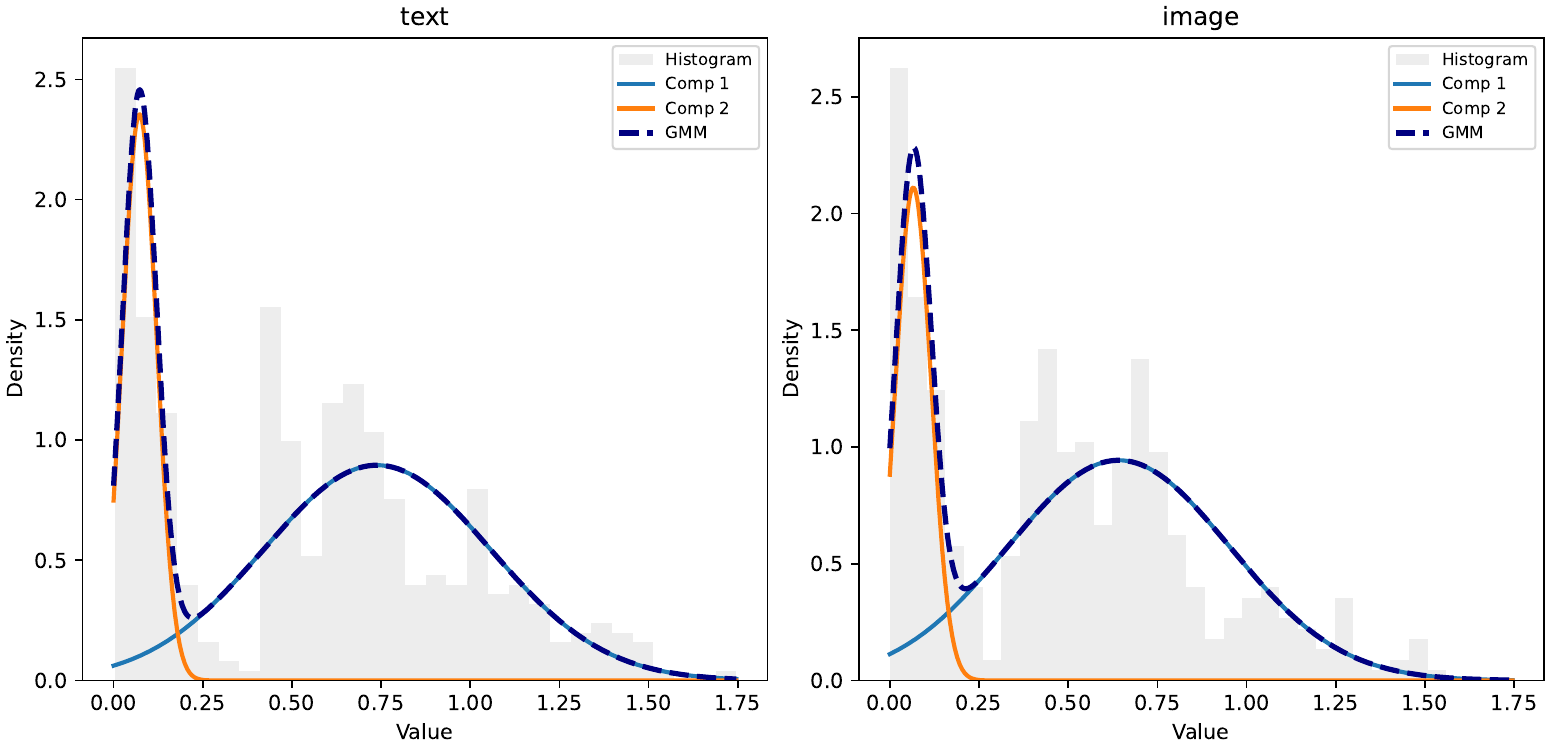}
        \label{fig:noise-added}
    }
    \caption{GMM fitting results in SAD on the CUB dataset without noise and with Gaussian noise of \(\sigma = 5\)}
    \label{fig:gmm}
\end{figure}

\subsubsection{Residual Cross-Class Reconstruction}
To further demonstrate that Residual Cross-Class Reconstruction (RCCR) effectively removes inter-class confusion, we visualize the representations learned with and without RCCR using t-SNE. In the version without RCCR, only the intra-class reconstruction loss in Eq. (\ref{eq:l-re}) is retained, while the cross-class residual reconstruction loss is removed. The results are shown in Fig. \ref{fig:tsne}, where the original input features of each modality are also visualized for comparison. It can be observed that, compared with the original input features in Fig. \ref{fig:tsne-ori}, the representations learned by the encoder without RCCR (Fig. \ref{fig:tsne-wo-cad}) exhibit improved intra-class compactness, indicating that the encoder captures class-relevant information. However, inter-class separability remains limited, suggesting the presence of inter-class confusion. In contrast, the representations learned with RCCR (Fig. \ref{fig:tsne-cad}) not only achieve stronger intra-class compactness but also demonstrate pronounced inter-class separability. This indicates that, with the help of RCCR, inter-class confusion is effectively removed from the learned representations, retaining only the most discriminative class-specific features. Additionally, some clusters contain a few samples from other classes, reflecting the presence of sample-level confusion.

\subsection{Discussion on sample-adaptive deconfusion}
The sample-adaptive deconfusion module consists of two components: confusion-free modality prior construction and sample-adaptive cross-modality rectification. Extensive experiments are conducted to further analyze the contribution of each component.
\subsubsection{Confusion-Free Modality Prior Construction}
The confusion-free modality prior construction provides reliable modality distribution priors to support the removal of confusion information within each sample. Therefore, experiments are conducted to evaluate its stability under noisy conditions. Specifically, Gaussian noise with $\epsilon=5$ is added to the image modality of the CUB dataset, and Gaussian mixture models (GMMs) fitted before and after noise addition are visualized. As shown in Fig. \ref{fig:gmm}, the histograms exhibit a rather clear bimodal quality distribution, indicating that the data can be effectively modeled using a two-component GMM. Moreover, the fitted GMM curves before and after noise addition show no significant differences, and the low-entropy peaks in both cases are close to 0. This is mainly because CAD has already removed most of the global noise, and the remaining sample-level confusion information has little impact on the overall fitting results. These observations demonstrate that even under noisy conditions, the confusion-free modality prior construction strategy can still select a sufficient number of high-confidence samples to obtain stable and reliable modality distribution priors.

\begin{figure}[t]
\centering
\includegraphics[width=0.95\columnwidth]{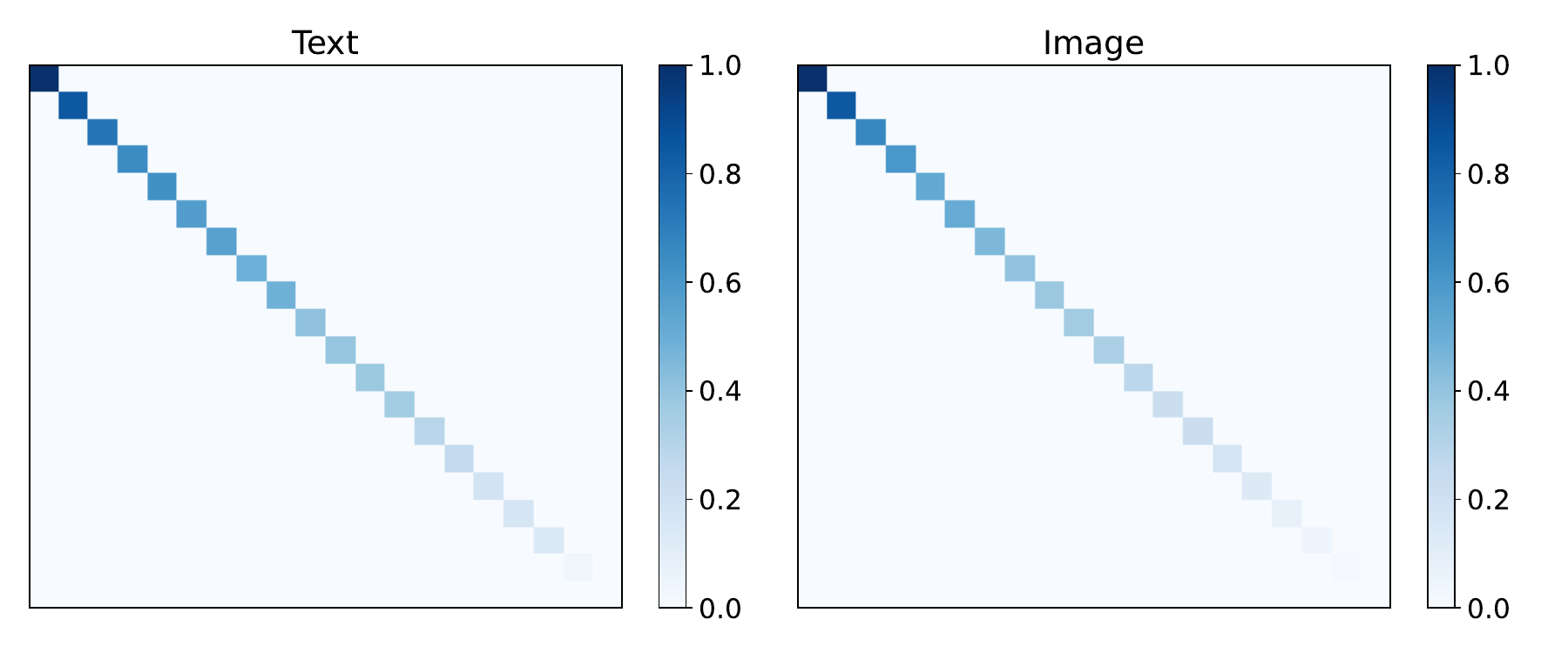} 
\caption{Visualization of the variances along the principal axes of the confusion-free prior covariances for each modality on the CUB dataset. For clarity, all variance values are normalized to the [0,1] range, and the variances of first 20 principal axes are shown.}
\label{fig:vis-variance-axes}
\end{figure}

\begin{figure}[t]
\centering
\includegraphics[width=0.95\columnwidth]{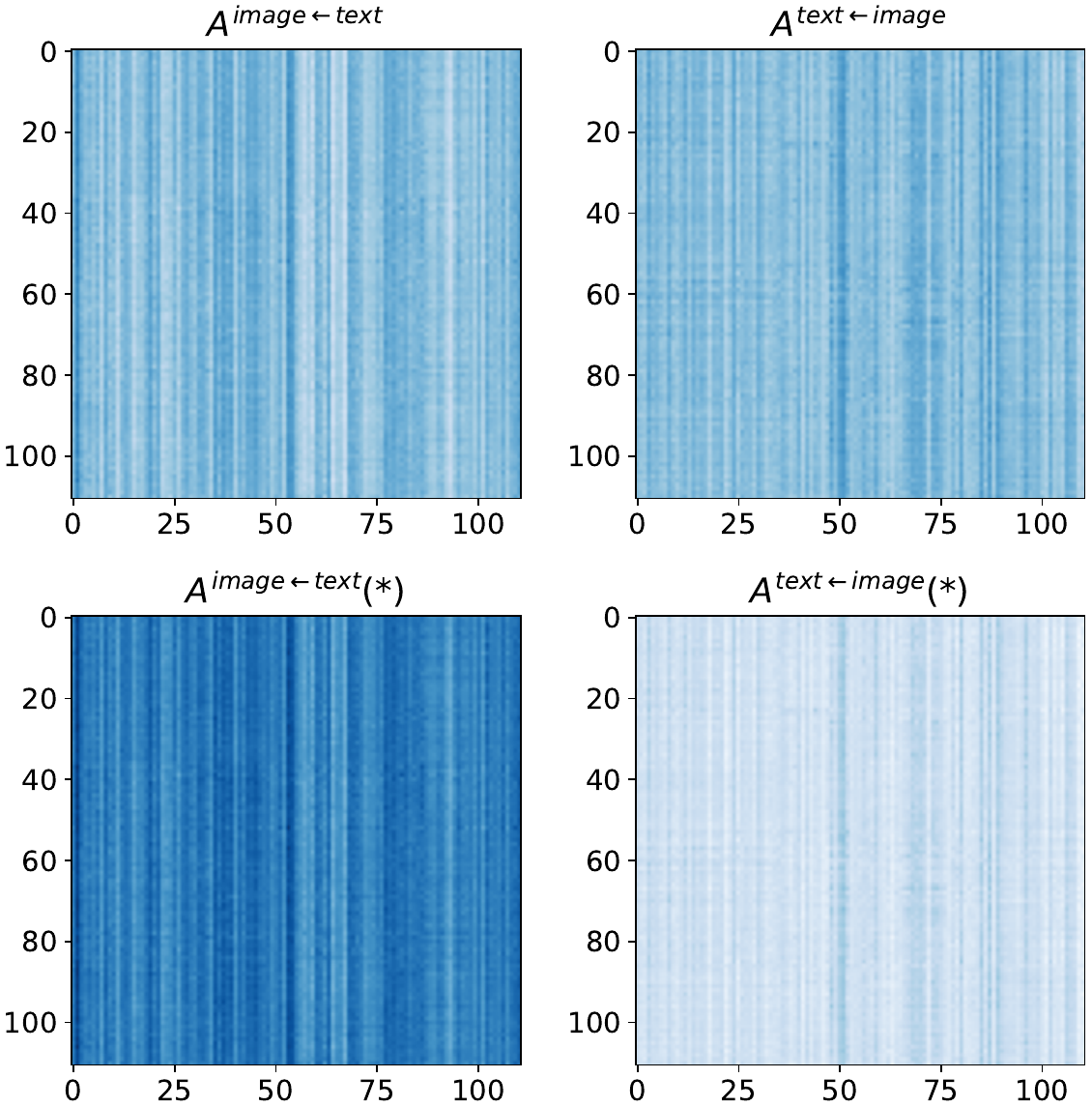} 
\caption{Comparison of the cross-modal attention maps \(A^{\text{image}\leftarrow \text{text}}\) and \(A^{\text{text}\leftarrow \text{image}}\) of a noise-affected sample of the CUB dataset before and after adding Gaussian noise with \(\sigma = 5\), where (*) denotes the results after noise corruption.}
\label{fig:attention-map}
\end{figure}

\subsubsection{Sample-Adaptive Cross-Modality Rectification}
Further experiments are conducted to investigate the role of Sample-Adaptive Cross-Modality Rectification in removing the confusion information of each modality within samples. Specifically, Gaussian noise with $\epsilon=5$ is sampled and added to the image modality of several randomly selected samples. Fig. \ref{fig:attention-map} shows the changes in the attention maps $A^{\text{image}\leftarrow\text{text}}$ and $A^{\text{text}\leftarrow\text{image}}$ on the CUB dataset that represent how the image modality obtains compensatory information from the text modality and vice versa of a sample, before and after noise corruption. After adding noise to the image data, the values in attention map $A^{\text{image}\leftarrow\text{text}}$ for information flow from text to image increase, while those from image to text ($A^{\text{text}\leftarrow\text{image}}$) decrease. This indicates that the cross-modality rectification mechanism effectively enables the lower-confidence modality to receive more complementary information from the higher-confidence modality, while suppressing information flow in the opposite direction. Therefore, the reliability of the rectification procedure is enhanced.

To further investigate the necessity of reweighting the compensatory features, we first visualize the variances \(\Lambda\) of each principal axes of the covariance of each modality prior \(p^m\). As shown in Fig. \ref{fig:vis-variance-axes}, the axes exhibit substantially different variance values, indicating that the degree of inter-class confusion varies across principal directions. In addition, we validate the effectiveness of the proposed reweighting strategy by removing the reweighting operation and by applying a negative reweighting variant. The ``w/o reweight'' setting directly adds the compensatory feature \(\Tilde{z}_i^m\) to the corresponding modality representation \(z^m_i\). The negative reweighting variant replaces the weight vector \(w=[w_1,\dots,w_{d^m}]\) with one positively correlated with the variances \(\lambda_1,\dots,\lambda_{d^m}\), i.e., replacing \(-\lambda_i\) with \(\lambda_i\) in Eq. (\ref{eq:weight}). As shown in Fig. \ref{fig:reweight}, across all datasets, reweighting yields the best performance, negative reweighting performs the worst, and w/o reweight lies in between. These results not only demonstrate the performance benefits of reweighting but also verify the necessity of assigning weights negatively correlated with the variances.

\begin{figure}[t]
\centering
\includegraphics[width=0.95\columnwidth]{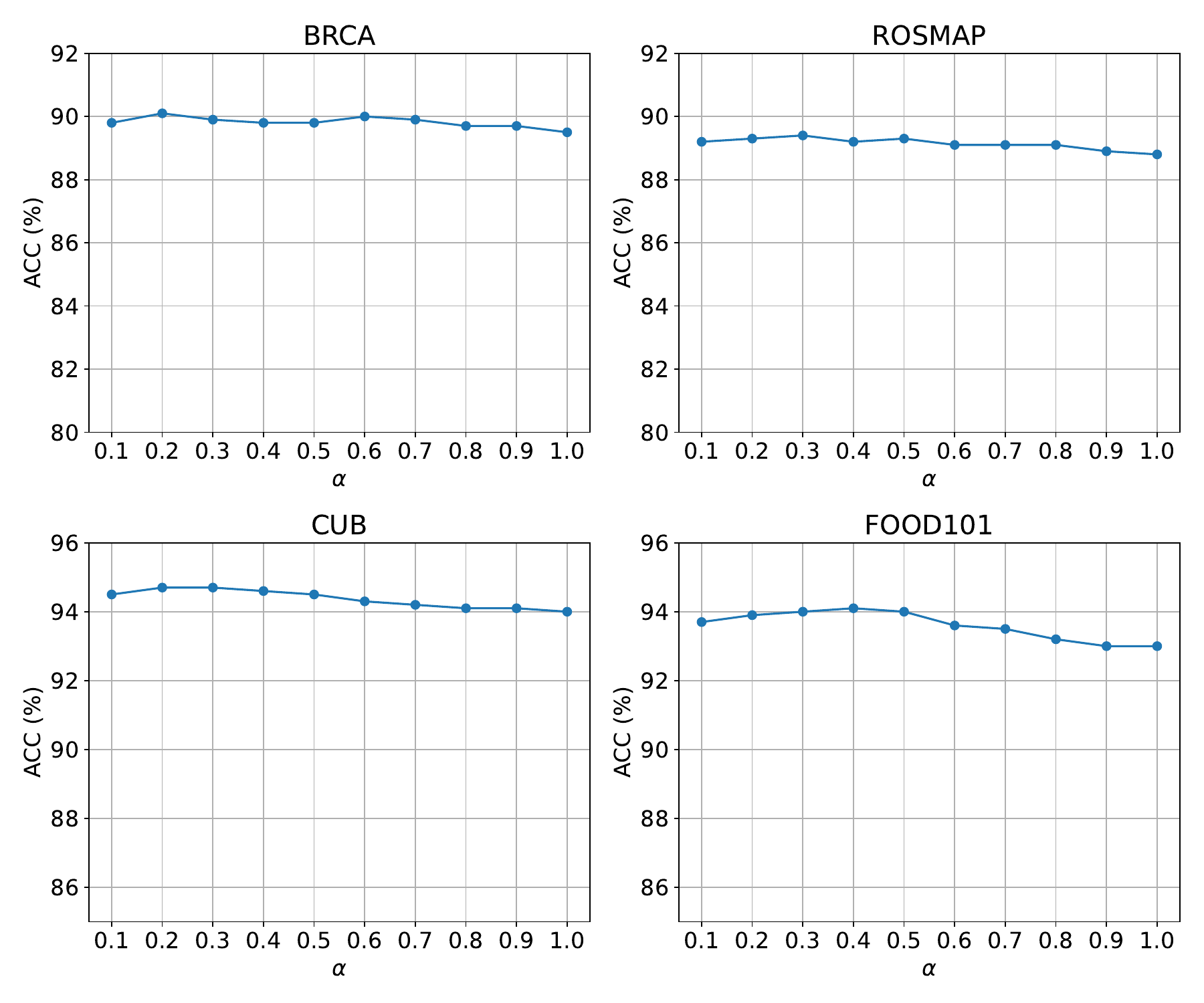} 
\caption{Analysis of the hyperparameter \(\alpha\) on the BRCA, ROSMAP, CUB, and FOOD101 datasets.}
\label{fig:param-analysis}
\end{figure}

\subsection{Parameter Analysis}
In the computation of the noise standard deviation \(\sigma_{c,m}\) in Eq. (\ref{eq:residual-cal}), a hyperparameter \(\alpha\) is introduced. Parameter analysis experiments are conducted to examine the effect of different \(\alpha\) values on model performance. As shown in Fig. \ref{fig:param-analysis}, variations in \(\alpha\) have only a minor impact on performance across all datasets, indicating that MLAD is relatively insensitive to this hyperparameter and exhibits strong robustness.

\section{Conclusion}
This paper observes that existing multimodal learning methods often learn features that are class-relevant but not sufficiently discriminative, which limits output confidence, particularly under noisy or low-quality data conditions. To address this, we propose Multi-Level Adaptive Deconfusion (MLAD), which enhances the discriminativeness and classification confidence of learned representations by removing inter-class confusion present in the data. MLAD considers inter-class confusion at both global and sample levels and employs class-adaptive deconfusion and sample-adaptive deconfusion to remove them correspondingly. Class-adaptive deconfusion adapts the encoder output depth according to the distinguishing difficulty of each class and effectively removes global-level confusion patterns within each class via residual cross-class reconstruction. Sample-adaptive deconfusion, guided by the confusion-free modality priors, performs sample-specific cross-modality rectification, effectively eliminating sample-level confusion patterns. Experiments on multiple benchmarks demonstrate that MLAD outperforms other reliable multimodal learning methods in classification performance and robustness. Ablation and analysis studies further validate the effectiveness of each component in MLAD.

Although existing methods have achieved reliable results on classification tasks, relatively little work has investigated their applicability to regression and other machine learning tasks. We believe this is an important and promising direction for further study, and we plan to extend the proposed method to a broader range of tasks in future work.

\section*{Acknowledgment} 
The authors wish to acknowledge the strong support and valuable suggestions provided by Professor Jiwen Lu from the Department of Automation at Tsinghua University for this work.

\bibliographystyle{IEEEtran}
\bibliography{reference.bib}

\end{document}